\documentclass[10pt,twocolumn,letterpaper]{article}

\usepackage[pagenumbers]{style}
\usepackage{amssymb}
\usepackage{fontawesome}

\usepackage{pifont}
\newcommand{\cmark}{\ding{51}}
\newcommand{\xmark}{\ding{55}}

\definecolor{color}{rgb}{0.21,0.49,0.74}
\definecolor{cell}{RGB}{230, 237, 233}
\usepackage[pagebackref,breaklinks,colorlinks,allcolors=color]{hyperref}

\usepackage{multirow}
\usepackage{titletoc}
\usepackage{soul}

\title{GEAL: Generalizable 3D Affordance Learning with Cross-Modal Consistency}

\author{
    Dongyue Lu\quad Lingdong Kong\quad Tianxin Huang\quad Gim Hee Lee 
    \\[1ex]
    National University of Singapore
    \\
    \texttt{dongyue.lu@u.nus.edu} \quad \texttt{lingdong@comp.nus.edu.sg}
    \\
    \quad \texttt{\{huangtx, gimhee.lee\}@nus.edu.sg}
    \\[1ex]
    \faGithubAlt~\textbf{Code \& Demo:} \href{https://dylanorange.github.io/projects/geal}{dylanorange.github.io/projects/geal}\\
    \faDownload~\textbf{Dataset \& Benchmark:} \href{https://huggingface.co/datasets/dylanorange/geal}{huggingface.co/datasets/dylanorange/geal}
}

\begin{document}

\maketitle

\begin{abstract}
Identifying affordance regions on 3D objects from semantic cues is essential for robotics and human-machine interaction. However, existing 3D affordance learning methods struggle with generalization and robustness due to limited annotated data and a reliance on 3D backbones focused on geometric encoding, which often lack resilience to real-world noise and data corruption. We propose \textbf{GEAL}, a novel framework designed to enhance the generalization and robustness of 3D affordance learning by leveraging large-scale pre-trained 2D models. We employ a dual-branch architecture with Gaussian splatting to establish consistent mappings between 3D point clouds and 2D representations, enabling realistic 2D renderings from sparse point clouds. A granularity-adaptive fusion module and a 2D-3D consistency alignment module further strengthen cross-modal alignment and knowledge transfer, allowing the 3D branch to benefit from the rich semantics and generalization capacity of 2D models. To holistically assess the robustness, we introduce two new corruption-based benchmarks: PIAD-C and LASO-C. Extensive experiments on public datasets and our benchmarks show that GEAL consistently outperforms existing methods across seen and novel object categories, as well as corrupted data, demonstrating robust and adaptable affordance prediction under diverse conditions. Code and corruption datasets have been made publicly available\footnote{GitHub: \url{https://github.com/DylanOrange/geal}}.

\end{abstract}    
\section{Introduction}
\label{sec:intro}

\begin{figure}[t]
    \centering
    \includegraphics[width=\linewidth]{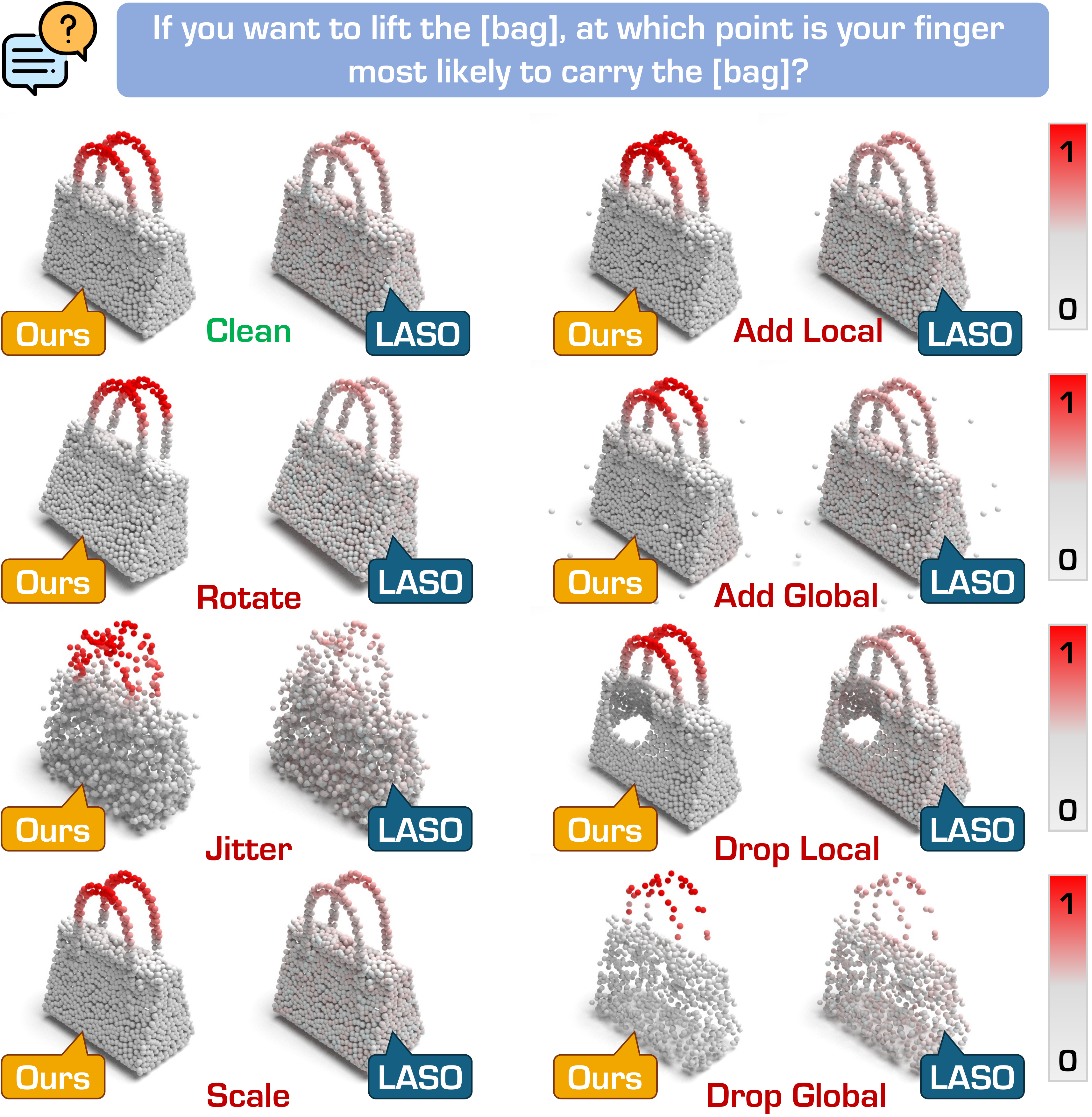}
    \vspace{-0.6cm}
    \caption{\textbf{3D affordance prediction under varied data noises.} Given a textual prompt, previous methods like LASO~\cite{li2024laso} (right side of each example) exhibit reduced robustness across different corruption types. In contrast, our proposed method, \textbf{GEAL} (left side of each example), maintains high accuracy and generalization across these challenging scenarios by effectively transferring knowledge from a large-scale pre-trained 2D foundation model, enhancing robustness and adaptability under diverse conditions.}
  \label{fig:teaser}
\end{figure}

3D affordance learning involves identifying interactive regions on objects given semantic cues such as image or textual instruction~\cite{cui2023strap, fang2018demo2vec}, which is a fundamental competency for intelligent systems~\cite{hou2021affordance, cruz2016training} to infer how an object can be used or manipulated~\cite{li2023locate, luo2022learning, yang2023grounding}. This understanding is vital for applications in robotics and human-machine interaction such as action prediction, object manipulation, and autonomous decision-making~\cite{gibson2014ecological, hassanin2021visual, geng2023rlafford, chuang2018learning}. For example, a robot equipped with affordance knowledge can intelligently interact with objects in its environment by determining where to grasp a handle or press a button.

Despite its potential, 3D affordance learning still faces significant challenges. Due to limited annotated data, 3D affordance models generally show poorer generalization than their 2D counterparts which benefit from abundant labeled data and large-scale pretraining~\cite{li2024one}. Additionally, 3D models often rely on backbones that focus on positional and geometric encoding, limiting their capacity to capture global semantic content and making them vulnerable to noisy or corrupted data from sensor inaccuracies, scene complexity, or processing artifacts in real-world settings \cite{ren2022pointcloud-c, kong2023robo3d, ren2022benchmarking, xie2024benchmarking, liu2023seal}. These issues further hinder the robustness and adaptability of current 3D affordance learning methods.

In this paper, we introduce a novel framework \textbf{GEAL}, which is designed to enhance the generalization and robustness of 3D affordance learning through a dual-branch architecture that leverages the correspondence between 2D and 3D data. GEAL generates realistic 2D renderings directly from sparse 3D data by employing 3D Gaussian splatting (3DGS) \cite{kerbl20233d} to build consistent mappings between 3D point clouds and 2D representations. This approach effectively creates a 2D branch from purely 3D data, which allows us to utilize the generalization capabilities and rich semantic knowledge of large-scale pre-trained 2D foundation models~\cite{radford2021learning,oquab2023dinov2} to enhance 3D affordance predictions.

We further introduce a granularity-adaptive fusion module, and a 2D-3D consistency alignment module to ensure robust multi-modal alignment. The granularity-adaptive fusion module dynamically integrates multi-level visual and textual features to address affordance queries at various scales and granularities. The 2D-3D consistency alignment module concurrently establishes reliable cross-modal correspondence with feature embeddings augmented to the Gaussian primitives of 3DGS, fostering effective knowledge transfer across branches, and enhancing the generalization and robustness of the 3D branch by enforcing consistent alignment between 2D and 3D modalities.

In view of the limitation of data scarcity to benchmark the robustness of 3D affordance models, we create two datasets: \textbf{PIAD-C}orrupt and \textbf{LASO-C}orrupt from existing commonly used affordance datasets~\cite{li2024laso, yang2023grounding}.
We design these benchmark datasets by incorporating various types of real-world corruptions such as scaling, cropping, \etc, to ensure their suitability in evaluating the robustness of 3D affordance models. By contributing these benchmark datasets, we aim to fill a critical gap in the affordance learning community by providing a standard for evaluating the robustness of point cloud-based 3D affordance methods. ~\cref{fig:teaser} shows an example of the text description and the corresponding 3D affordance on 3D point clouds that are corrupted under various noise types.

We validate the generalization and robustness of our GEAL on both standard and corruption-based benchmarks, demonstrating that our approach consistently outperforms recent methods in all scenarios. Our experiments confirm that our GEAL effectively transfers knowledge from seen to unseen data and maintains high performance even under corruption, underscoring the 
adaptability of our framework across challenging scenarios.

In summary, the main contributions of this work can be summarized as follows: 
\begin{itemize} 
    \item We propose \textbf{GEAL}, a novel approach for generalizable 3D affordance learning. By employing 
    3DGS, we develop a 2D affordance prediction branch for 3D point clouds, harnessing the robust generalization and semantic understanding of pre-trained 2D foundation models.
    
    \item We propose granularity-adaptive fusion and 2D-3D consistency alignment to integrate and propagate knowledge across the dual-branch architecture, and enhance the generalizability of the 3D branch using 2D knowledge. 
    
    \item We establish two corruption-based benchmarks: \textit{PIAD-C} and \textit{LASO-C}, to holistically evaluate the robustness of 3D affordance learning under real-world scenarios, contributing a standard to the community for robustness analysis. 
    
    \item Extensive experiments validate the strong performance of our approach on both mainstream and corruption 3D affordance learning benchmarks, proving its generalization ability and robustness across diverse conditions. 
\end{itemize}
\section{Related Work}
\label{sec:relatedwork}
\noindent \textbf{2D Affordance Learning.} Affordances refer to potential actions that objects or environments enable for an observer, based on their properties\cite{koppula2013learning, do2018affordancenet, chuang2018learning, myers2015affordance}. Early methods for affordance detection mainly try to identify interaction regions in images and videos \cite{Li2023G2L, luo2021one, roy2016multi, thermos2020deep, zhao2020object}, though these often lacked precise localization of affordance-relevant object parts. To address this, later research improved affordance localization \cite{fang2018demo2vec, lu2022phrase, luo2022learning, luo2022grounded, nagarajan2019grounded, li2023locate, yang2023grounding, chen2023affordance, liu2022joint} given demonstration 2D data. Recently, large-scale pre-trained models~\cite{radford2021learning, caron2021emerging} have aligned visual features with affordance-related textual descriptions, reducing dependence on manual labels and enhancing affordance prediction in new contexts~\cite{lu2022phrase, mi2020intention, mi2019object, nguyen2023open}. Building on this, some studies~\cite{li2024one, li2023locate, hadjivelichkov2023one} turn to leverage foundation models to generalize affordance detection to novel objects and views.

\vspace{2mm}
\noindent \textbf{3D Affordance Learning.} Extending affordance detection to 3D space presents challenges due to the need for accurate spatial and depth information. While some studies use 2D data to detect 3D affordance regions \cite{chuang2018learning, do2018affordancenet, Li2023G2L}, they often struggle with precise 3D interaction sites. The availability of large-scale 3D object datasets \cite{Geng_2023_CVPR, Liu_2022_CVPR, Mo_2019_CVPR} has driven efforts to map affordances directly onto 3D structures \cite{deng20213d, mo2022o2o, xu2022partafford, li2024laso, yang2023grounding}, aiming to capture complex spatial relationships. Recent methods~\cite{nguyen2023open, huang2023voxposer, li2024laso} leverage 2D visual and language models for open-vocabulary affordance detection, enhancing generalization without fixed label sets. Despite these advancements, achieving robust generalization in 3D remains challenging, as 3D backbones still lack the generalization capabilities of 2D foundation models, thus, our method leverages large-scale 2D foundation models to improve 3D affordance generalization.

\begin{figure*}[t]
    \centering
    \includegraphics[width=\linewidth]{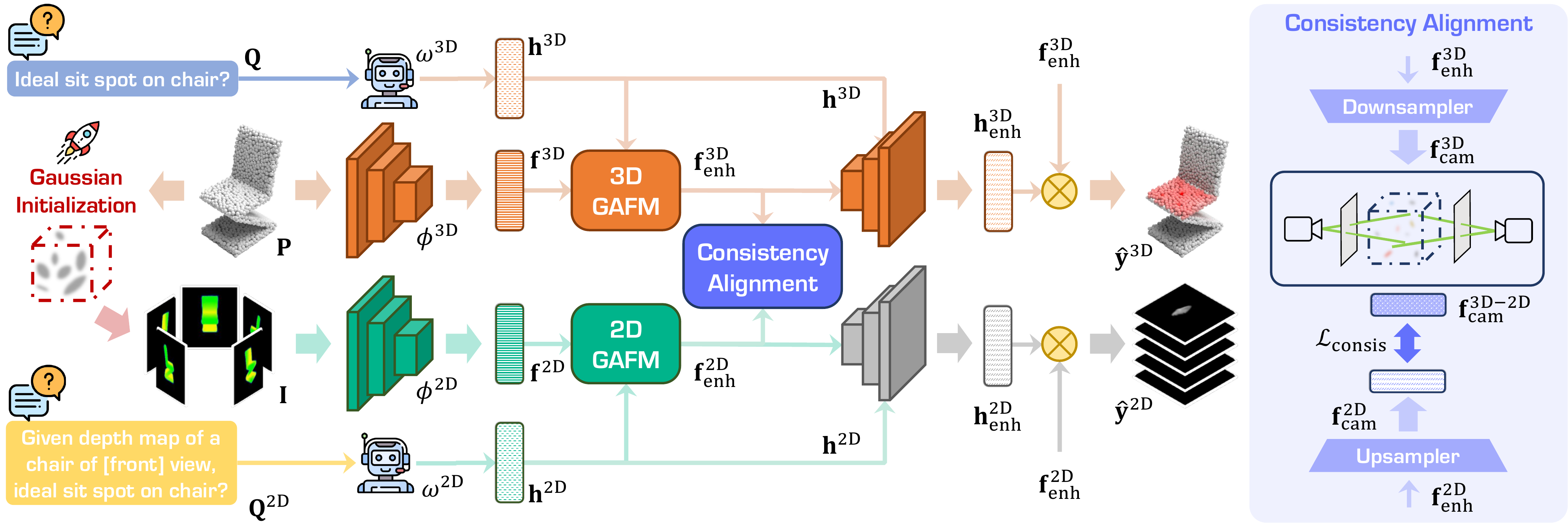}
    \vspace{-0.6cm}
    \caption{
    \textbf{(Left): Framework Overview.} The proposed \textbf{GEAL} consists of two branches: 3D and 2D. The 2D branch is established through 3D Gaussian Splatting to leverage the generalization capabilities of large pre-trained 2D models (\cf~\ref{sec:preliminaries}). We then perform cross-modality alignment, including \textbf{Granularity-Adaptive Visual-Textual Fusion} and \textbf{2D-3D Consistency Alignment}, to unify features from different modalities into a shared embedding space (\cf~\ref{sec:alignment}). Finally, we decode generalizable affordance from this embedding space (\cf~\ref{sec:decoding}).
    \textbf{(Right): Architecture of the 2D-3D Consistency Alignment Module.} This module maps features from 2D and 3D modalities into a shared embedding space and enforces consistency alignment to enable effective knowledge transfer across branches.
    }
  \label{fig:overview}
\end{figure*}

\vspace{2mm}
\noindent \textbf{Robustness for 3D Affordance Learning.}
Real-world 3D affordance learning faces inevitable challenges from point cloud corruptions caused by scene complexity, sensor inaccuracies, and processing errors~\cite{hendrycks2019benchmarking,ren2022benchmarking,xie2024benchmarking,li2024place3d}. Existing studies aim to improve and benchmark robustness against noise and corruption in 3D perception across real-world scenarios~\cite{wang2023sample,kim2021point, lee2021regularization, wang2024target, kong2023robo3d,kong2023robodepth,hao2024mapbench}. However, affordance learning uniquely requires precise identification of interactive regions under variable and degraded data conditions. To our knowledge, this work is the first to specifically address robustness in 3D affordance learning, providing a targeted approach to enhance reliability across diverse environments.
\section{Methodology}
\label{sec:method}

In this section, we describe the technical components of our proposed \textbf{GEAL} framework. 
{An overview of the full framework is shown in \cref{fig:overview}.}
Given an instruction \( Q \) and an object point cloud \( \mathbf{P} \in \mathbb{R}^{N \times 3} \) with \( N \) points,  GEAL predicts an affordance score \( \mathbf{y} \in \mathbb{R}^N \), where each value in \( \mathbf{y} \) indicates the likelihood that a corresponding point supports the specified functionality. In \cref{sec:preliminaries}, we employ Gaussian splatting as a cross-modal mapping to bridge {the} 2D and 3D modalities, establishing a 2D branch to leverage the generalization and robustness of large pre-trained 2D models. \cref{sec:alignment} details our cross-modal alignment strategy, incorporating both granularity-adaptive visual-text fusion and 2D-3D consistency alignment to unify these modalities in the embedding space. In \cref{sec:decoding}, we outline the decoding process that derives robust and generalizable affordance predictions from the aligned feature space. 

\subsection{3D-2D Mapping with Gaussian Splatting}
\label{sec:preliminaries}

\textbf{Motivation.} Current 3D affordance learning methods suffer from poor generalization due to limited annotated data and exhibit relatively weak robustness owing to limited global semantic capture. In contrast, 2D affordance learning methods \cite{li2024one, li2023locate} leverage 2D foundation models \cite{oquab2023dinov2, radford2021learning} pretrained on large amounts of data, offering superior generalization and robustness. 
A 3D-2D mapping to leverage {the} 
strengths {of 2D foundation models} is {thus} promising. 
{However, a direct projection of 3D point clouds onto 2D planes yields sparse 2D points without semantic and depth information that are not useful for feature extraction with 2D foundation models.}
To overcome this {issue}, we adopt 3D Gaussian Splatting \cite{kerbl20233d} which represents 3D scenes as learnable Gaussian primitives 
{for} realistic, differentiable and high-speed rendering from arbitrary viewpoints. This approach allows us to synthesize realistic 2D images from sparse point clouds, preserving crucial semantic and depth information for downstream affordance learning tasks. 
{Moreover, 3D Gaussian Splatting offers smoother transitions between points, preserves occlusions and depth cues for a coherent and accurate scene representation.}

\vspace{2mm}
\noindent\textbf{Gaussian Initialization.} In 3D Gaussian Splatting, each Gaussian primitive is characterized by its 3D position ${\mu}$ represented by a 3D coordinate, a covariance matrix ${\Sigma}$ which defines its shape and spread, spherical harmonic parameters ${c}$ representing its color, and an opacity value $\alpha$ that indicates its transparency. To render 3D Gaussian primitives into 2D image planes, we apply point-based $\alpha$-blending using a tile-based rasterizer for efficient rendering. The rendered color at each pixel $v$ is calculated as follows:
\begin{equation}
    C(v) = \sum\nolimits_{i \in \mathcal{N}} c_i \alpha_i \prod\nolimits_{j=1}^{i-1} (1 - \alpha_j),
    \label{eq:rendering_3dgs}
\end{equation}
where $c_i$ is the color of the $i$-th Gaussian, $\mathcal{N}$ represents the Gaussians within the tile, and $\alpha_i = o_i G^{2D}_i(v)$. 
$o_i$ is opacity of the $i$-th Gaussian and $G^{2D}_i(\cdot)$ represents the function of the $i$-th Gaussian projected onto 2D. Similarly, a depth map can be rendered as:
\begin{equation}
    D(v) = \sum\nolimits_{i \in \mathcal{N}} d_i \alpha_i \prod\nolimits_{j=1}^{i-1} (1 - \alpha_j),
    \label{eq:rendering_depth_3dgs}
\end{equation}
where $d_i$ denotes the depth of the $i$-th Gaussian primitive under {the} provided camera pose.


To ensure that the rendered images accurately reflect the geometry of the input point cloud \( \mathbf{P} \), we set the Gaussian mean positions to match the point coordinates, \ie \( \boldsymbol{\mu} = \mathbf{P} \). The covariance \( \boldsymbol{\Sigma} \) and opacity \( \boldsymbol{\alpha} \) are manually adjusted and then {kept} fixed {during training} to preserve the original geometry. 
Using the depth map from Eq.~\eqref{eq:rendering_depth_3dgs} with \( V \) camera poses and a predefined color map, we obtain realistic images \( \mathbf{I} \in \mathbb{R}^{V \times 3 \times H \times W} \) that preserve both semantics and spatial information of the original point cloud, effectively bridging the 3D-2D gap. Treating the affordance score \( \mathbf{y} \in [0, 1] \) as grayscale color, we assign 
{the color of each Gaussian} to match its affordance score, \ie \( \mathbf{c} = \mathbf{y} \). We generate 2D affordance masks \( \mathbf{y}_{\text{2D}} \in \mathbb{R}^{V \times H \times W} \), where each pixel represents the affordance score of the associated 3D point. This process establishes a coherent mapping from 3D point clouds and affordance scores to their 2D counterparts, using Gaussian splatting to generate realistic, informative 2D representations that enhance affordance learning.

\vspace{2mm}
\noindent\textbf{Encoding.} Our \textbf{GEAL} framework {as shown in} \cref{fig:overview} comprises {a} 2D and {a} 3D branch with backbones \( \phi^{\text{2D}}(\cdot) \) and \( \phi^{\text{3D}}(\cdot) \), respectively. The 3D branch uses PointNet++ \citep{qi2017pointnet++} for point cloud feature extraction, while the 2D branch employs DINOV2 \citep{oquab2023dinov2} for image features. Both networks produce multi-scale features at various granularities.
At each scale \( i \), we extract features:
\begin{equation}
    \mathbf{f}_{i}^{\text{3D}} = \phi_{i}^{\text{3D}}(\mathbf{P}), \quad
    \mathbf{f}_{i}^{\text{2D}} = \phi_{i}^{\text{2D}}(\mathbf{I}),
\end{equation}
where \( \mathbf{f}_{i}^{\text{3D}} \in \mathbb{R}^{B \times C_{i}^{\text{3D}} \times N_{i}^{P}} \) and \( \mathbf{f}_{i}^{\text{2D}} \in \mathbb{R}^{B \times V \times C^{\text{2D}} \times N^{I}} \). 
\( B \) is the batch size, \( V \) is the number of views, and \( C_{i}^{\text{3D}} \) and \( C^{\text{2D}} \) are feature dimensions. $N_{i}^P$ is the number of point in scale $i$, and $N^{I}$ is image patch length. 
Note that the 3D spatial resolution $N_i^P$ and $C_i^{3D}$ differ between different scales due to the usage of PointNet++ \cite{qi2017pointnet++}.

We process the input prompt \( Q \) using lightweight language models \( \omega^{\text{3D}}(\cdot) \) and \( \omega^{\text{2D}}(\cdot) \) 
{that share} the same architecture. For the 2D input, we modify the prompt to \( Q^{\text{2D}} \) by adding: ``\textbf{\textit{Given a depth map of a} [object] \textit{in} [view]}", constructing \textbf{view-dependent prompt} to enhance context understanding. The text embeddings are:
\begin{equation}
    \mathbf{h}^{\text{3D}} = \omega^{\text{3D}}(Q), \quad
    \mathbf{h}^{\text{2D}} = \omega^{\text{2D}}(Q^{\text{2D}}),
\end{equation}
where \( \mathbf{h}^{\text{3D}} \in \mathbb{R}^{B \times C^{\text{txt}} \times L} \) and \( \mathbf{h}^{\text{2D}} \in \mathbb{R}^{B \times V \times C^{\text{txt}} \times L} \), with \( L \) as the sequence length.

\subsection{Cross-Modal Consistency Alignment}
\label{sec:alignment}
Since point cloud, image, and text features are embedded in distinct spaces, we design alignment modules to map these multi-modal features into a shared embedding space. First, we fuse visual features at varying granularities with textual features through Granularity-Adaptive Visual-Textual Fusion, supporting affordance learning conditioned on instructions across different scales. 
{Subsequently}, we propagate knowledge from the 2D to 3D branch via a 2D-3D Consistency Alignment by enforcing consistency between 2D and 3D features.

\begin{figure}[t]
    \centering
    \includegraphics[width=\linewidth]{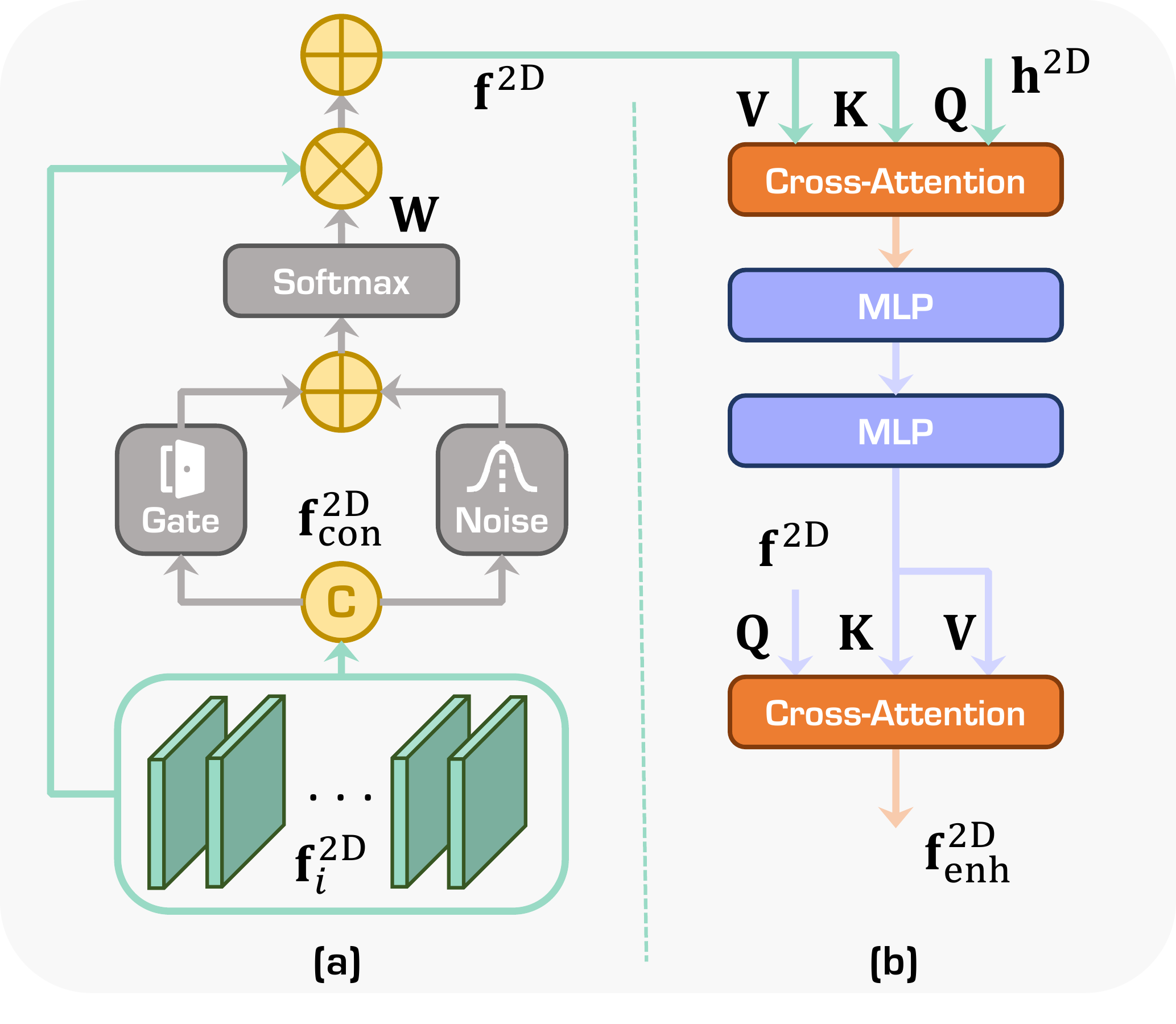}
    \vspace{-0.6cm}
    \caption{
        {Illustration of the \textbf{Granularity-Adaptive Fusion Module}, it consists of a Flexible Granularity Feature Aggregation mechanism (a) and a Text-Conditioned Visual Alignment mechanism (b), we take the 2D branch as an example.} 
    }
  \label{fig:module}
\end{figure}

\vspace{2mm}
\noindent\textbf{Granularity-Adaptive Visual-Textual Fusion.}  Both 2D and 3D backbones capture different levels of granularity, with lower layers focusing on fine details and higher layers providing broader context. Since affordances can span multiple object parts, leveraging features at various granularities is advantageous. To achieve this, we introduce \textbf{G}ranularity-\textbf{A}daptive \textbf{F}usion \textbf{M}odule \textbf{(GAFM)}, which integrates multi-granularity visual features with textual cues via {\textit{Flexible Granularity Feature Aggregation}} and \textit{Text-Conditioned Visual Alignment}. These mechanisms allow the model to adaptively fuse features across different granularities, enhancing affordance prediction in response to specific instructions. An illustration of the Granularity-Adaptive Fusion Module is provided in ~\cref{fig:module}.

\noindent \textit{\textbf{Flexible Granularity Feature Aggregation.}} This mechanism aims to fuse visual features from different granularities.
Taking the 2D branch as an example, we concatenate feature maps from the last \( m \) levels, forming an input tensor \( \mathbf{f}_{\text{con}}^{\text{2D}} \in \mathbb{R}^{ B \times V \times (m \times C^{\text{2D}}) \times N^{I}} \). 
Inspired by previous works \cite{li2024one,zhong2024convolution}, we {then} compute adaptive soft weights to regulate the contribution of each feature level, enabling the model to adapt to varying levels of detail. These weights are computed via a gating function with learned noise, introducing perturbations that enhance adaptability:
\begin{equation}
    \mathbf{W} = \operatorname{Softmax} \left( \mathbf{f}_{\text{con}}^{\text{2D}} \cdot \mathbf{W}_{g} + \sigma \cdot \epsilon \right),
    \label{eq:soft_weight}
\end{equation}
where \( \mathbf{W}_{g} \in \mathbb{R}^{(m \times C^{\text{2D}}) \times m} \) is a trainable weight matrix, \( \mathbf{W} \in \mathbb{R}^{B \times m} \) represents the concatenation of weights \( w_i \) for each feature level, \( \sigma \) is a learned standard deviation controlling the noise scale, and \( \epsilon \sim \mathcal{N}(0, 1) \) is Gaussian noise. These weights balance the influence of each feature level, enabling affordance reasoning across different granularities.

The fused feature map is then obtained by applying the adaptive weights to 
features {from each level}:
\begin{equation}
    \mathbf{f}^{\text{2D}} = \sum\nolimits_{i=1}^{m} w_{i} \odot \mathbf{f}_{i}^{\text{2D}},
\end{equation}
where \( \odot \) denotes element-wise multiplication {and} $w_{i} \in \mathbf{W}$. This adaptive aggregation yields a robust feature representation across varying conditions 
{to enhance the generalization ability of the model.} 

\noindent \textit{\textbf{Text-Conditioned Visual Alignment.}} This module is proposed to integrate visual features with the textual instruction.
we {follow \cite{tolstikhin2021mlp, li2024laso} to} feed $\mathbf{f}^{\text{2D}}$ and $\mathbf{h}^{\text{2D}}$ into a transformer block. We first enhance the textual features $\mathbf{h}^{\text{2D}}$ with visual features $\mathbf{f}^{\text{2D}}$ through cross-attention, followed by refinement with two multilayer perceptrons (MLPs). 
We then acquire the visual features $\mathbf{f}^{\text{2D}}_{\text{enh}} \in \mathbb{R}^{B \times C^{\text{txt}} \times N^{I}}$ 
by querying the refined textual features with cross-attention. 
We {thus} ensure that the 2D visual features maintain their spatial structure while embedding the question-relevant information.

In the 3D branch, we align textual features with multi-granularity visual features in a similar manner. However, due to varying spatial resolutions and feature dimensions across scales in PointNet++ \cite{qi2017pointnet++}, directly concatenating all scales' features and computing soft weights as in ~\cref{eq:soft_weight} is not feasible. To address this, we first apply Text-Conditioned Visual Alignment to the 3D visual features at each scale, then upsample them to a uniform resolution. Finally, we perform Flexible Granularity Feature Aggregation on these upsampled features to produce the aggregated visual representation.

\vspace{2mm}
\noindent\textbf{2D-3D Consistency Alignment.}  
2D features retain rich semantic context and strong generalization via the pre-trained backbone \cite{oquab2023dinov2}, while 3D features preserve geometric and spatial details, compensating for the loss of 2D information caused by self-occlusions. To propagate the knowledge inherently, we introduce \textbf{C}onsistency \textbf{A}lignment \textbf{M}odule \textbf{(CAM)} to ensure mutual alignment and knowledge transfer from 2D to 3D spaces.

Specifically, as shown in right part of \cref{fig:overview}, we map \( \mathbf{f}_{\text{enh}}^{\text{3D}} \) and \( \mathbf{f}_{\text{enh}}^{\text{2D}} \) into a shared embedding space. Given the 2D-3D correspondence, regions in 2D and 3D representations that map to the same spatial areas should exhibit similar feature representations. 
By enforcing this consistency, we facilitate 2D-3D knowledge propagation to enhance the understanding of affordances across both modalities of the model.

To align 3D and 2D features in the same embedding space, we employ a down-sampler consisting of two Conv1D layers 
{that reduces} the feature dimension of \( \mathbf{f}_{\text{enh}}^{\text{3D}} \) to $\mathbf{f}_{\text{cam}}^{\text{3D}} \in \mathbb{R}^{B \times C^{\text{cam}} \times N}$. This processed feature acts as the representation for each point.
%
We {then} leverage the established 2D-3D mapping using Gaussian splatting {to project these point features into 2D}. For each Gaussian, we treat its point feature vector as an inherent attribute. The 2D feature at pixel \( v \) is then rendered as:
\begin{equation}
    \mathbf{F}(v) = \sum_{i \in \mathcal{N}} \mathbf{f}_i \alpha_i \prod_{j=1}^{i-1} (1 - \alpha_j),
    \label{eq:rendering_feature_3dgs}
\end{equation}
where \( \mathbf{f}_i \) is the feature of the \( i \)-th Gaussian, \( \alpha_i \) is its opacity, and \( \mathbf{F}(v) \) is the resulting semantic feature at pixel \( v \).

Similarly, we map the 2D features into the same embedding space using an up-sampler consisting of three Conv2D layers, which upsamples the spatial resolution of \( \mathbf{f}_{\text{enh}}^{\text{2D}} \) while also reducing its feature dimension to \( \mathbf{f}_{\text{cam}}^{\text{2D}} \in \mathbb{R}^{B \times V \times C^{\text{cam}} \times H \times W} \). 
\( V \) 
{is} the number of views and \( H \) and \( W \) 
{is} the spatial dimensions. Given the pixel positions from all V number of $H*M$ feature maps as $M$, we can define the 3D-2D projected feature as $\mathbf{f}_{\text{cam}}^{\text{3D}-{\text{2D}}}=\{\mathbf{F}(v)|v \in M\}$.
We then enforce a consistency constraint by minimizing the difference between the aligned 3D-2D features using \( L_2 \) loss:
\begin{equation}
    \mathcal{L}_{\text{consis}} = \text{MSE}(\mathbf{f}_{\text{cam}}^{\text{3D}-{\text{2D}}},~ \mathbf{f}_{\text{cam}}^{\text{2D}}).
\end{equation}
This consistency loss \( \mathcal{L}_{\text{consis}} \) encourages the model to maintain similar representations in both 2D and 3D spaces, effectively aligning affordance knowledge across domains. This alignment supports 2D-3D knowledge propagation, ensuring that {the} information learned in the 2D domain benefits the 3D features. 
 
\subsection{Decoding Generalizable Affordance}
\label{sec:decoding}
The affordance scores is decoded under the condition of affordance instructions. Through transformer decoder, the textual features attend to enhanced visual features, focusing the model on specific object parts for accurate predictions.

Our decoder architecture is shared across both 2D and 3D branches. In the 2D branch, textual features \( \mathbf{h}^{\text{2D}} \) and enhanced visual features \( \mathbf{f}_{\text{enh}}^{\text{2D}} \) are processed through a 3-layer transformer decoder. Here, \( \mathbf{h}^{\text{2D}} \) serves as the query, and \( \mathbf{f}_{\text{enh}}^{\text{2D}} \) acts as key and value, yielding updated textual features \( \mathbf{h}_{\text{enh}}^{\text{2D}} \). Each layer comprises self-attention to refine textual relationships and cross-attention to guide focus toward relevant visual regions.

After the transformer decoder, these enhanced textual features serve as dynamic kernels to predict affordance scores from visual features. The final affordance prediction \( \hat{\mathbf{y}}^{\text{2D}} \) is obtained by multiplying \( \mathbf{h}_{\text{enh}}^{\text{2D}} \) with \( \mathbf{f}_{\text{enh}}^{\text{2D}} \), followed by a sigmoid activation:
\begin{equation}
    \hat{\mathbf{y}}^{\text{2D}} = \text{sigmoid}\left( \mathbf{h}_{\text{enh}}^{\text{2D}} \cdot \mathbf{f}_{\text{enh}}^{\text{2D}} \right),
\end{equation}
where \( \hat{\mathbf{y}}^{\text{2D}} \in \mathbb{R}^{N} \) denotes affordance scores.

\vspace{2mm}
\noindent\textbf{Training.} 
We employ a combination of Binary Cross-Entropy (BCE) and Dice loss to guide the affordance score prediction in each branch, addressing both class imbalance and segmentation accuracy. For the 2D branch, the loss function is:
\begin{equation}
    \mathcal{L}^{\text{2D}} = \mathcal{L}_{\text{BCE}}^{\text{2D}} + {\mathcal{L}_{\text{Dice}}^{\text{2D}}},
\end{equation}
where \( \mathcal{L}_{\text{BCE}}^{\text{2D}} \) minimizes discrepancies between predicted and true affordance scores, and \( \mathcal{L}_{\text{Dice}}^{\text{2D}} \) improves the overlap between predicted and ground truth regions by maximizing intersection over union. 

We adopt a two-stage training approach. 
We train the 2D branch {in the first stage}, optimizing it for robust feature extraction and affordance decoding.
{Except for the CAM (Conisistency Alignment Module), all layers in the 2D branch are frozen in the second stage training.}
This approach allows the 3D branch to leverage fixed 2D features while adapting to 3D-specific characteristics. 
{Consequently}, the loss function for the 3D branch becomes:
\begin{equation}
    \mathcal{L}^{\text{3D}} = \mathcal{L}_{\text{BCE}}^{\text{3D}} + \mathcal{L}_{\text{Dice}}^{\text{3D}} + \mathcal{L}_{\text{consis}}.
\end{equation}

During inference, only the 3D branch is used, ensuring efficient and lightweight affordance prediction.

\subsection{Corrupt Data Benchmark}

To facilitate robust 3D affordance learning across diverse real-world scenarios, we establish two 3D affordance robustness benchmarks: \textbf{PIAD-C} and \textbf{LASO-C} based on the test sets of the commonly used datasets \textbf{PIAD} and \textbf{LASO} following \cite{ren2022pointcloud-c}. We apply seven types of corruptions -- $^1$\textit{Add Global}, $^2$\textit{Add Local}, $^3$\textit{Drop Global}, $^4$\textit{Drop Local}, $^5$\textit{Rotate}, $^6$\textit{Scale}, and $^7$\textit{Jitter} -- each with five severity levels. This results in a total of $4,890$ object-affordance pairings, comprising $17$ affordance categories and $23$ object categories with $2,047$ distinct object shapes. More details are provided in the supplementary material.

\section{Experiments}
\label{sec:experiment}

\begin{table}[t]
    \centering
    \caption{The overall results of all comparative methods on \textbf{PIAD} \cite{yang2023grounding}. \textbf{Seen} and \textbf{Unseen} are two partitions of the dataset. AUC and aIOU are shown in percentage. The \textbf{best} and \underline{2nd best} scores from each metric are highlighted in \textbf{bold} and \underline{underlined}, respectively.}
    \vspace{-0.2cm}
    \resizebox{\linewidth}{!}{
    \begin{tabular}{c|c|cccc}
    \toprule
    \textbf{Type} & \textbf{Method} & \textbf{aIoU} $\uparrow$ & \textbf{AUC} $\uparrow$ & \textbf{SIM} $\uparrow$ & \textbf{MAE} $\downarrow$ 
    \\
    \midrule\midrule
    \multirow{9}{*}{\textbf{Seen}} 
    & MBDF~\cite{tan2021mbdf} & $9.3$ & $74.9$ & $0.415$ & $0.143$ 
    \\
    & PMF~\cite{zhuang2021perception} & $10.1$ & $75.1$ & $0.425$ & $0.141$ 
    \\
    & FRCNN~\cite{xu2022fusionrcnn} & $12.0$ & $76.1$ & $0.429$ & $0.136$ 
    \\
    & ILN~\cite{chen2022imlovenet} & $11.5$ & $75.8$ & $0.427$ & $0.137$ 
    \\
    & PFusion~\cite{xu2018pointfusion} & $12.3$ & $77.5$ & $0.432$ & $0.135$ 
    \\
    & XMF~\cite{aiello2022cross} & $12.9$ & $78.2$ & $0.441$ & $0.127$ 
    \\ 
    & IAGNet~\cite{yang2023grounding} & \underline{$20.5$} & \underline{$84.9$} & $0.545$ & $0.098$ 
    \\
    & LASO~\cite{li2024laso} & $19.7$ & $84.2$ & \underline{$0.590$} & \underline{$0.096$} 
    \\
    & \cellcolor{cell}\textbf{GEAL (Ours)} & \cellcolor{cell}$\mathbf{22.5}$ & \cellcolor{cell}$\mathbf{85.0}$ & \cellcolor{cell}$\mathbf{0.600}$ & \cellcolor{cell}$\mathbf{0.092}$ 
    \\ 
    \midrule
    \multirow{9}{*}{\textbf{Unseen}} 
    & MBDF~\cite{tan2021mbdf} & $4.2$ & $58.2$ & $0.325$ & $0.213$ 
    \\
    & PMF~\cite{zhuang2021perception} & $4.7$ & $60.3$ & $0.330$ & $0.211$ 
    \\
    & FRCNN~\cite{xu2022fusionrcnn} & $5.1$ & $61.9$ & $0.332$ & $0.195$ 
    \\
    & ILN~\cite{chen2022imlovenet} & $4.7$ & $59.7$ & $0.325$ & $0.207$ 
    \\
    & PFusion~\cite{xu2018pointfusion} & $5.3$ & $61.9$ & $0.330$ & $0.193$ 
    \\
    & XMF~\cite{aiello2022cross} & $5.7$ & $62.6$ & $0.342$ & $0.188$ 
    \\ 
    & IAGNet~\cite{yang2023grounding} & \underline{$8.0$} & \underline{$71.8$} & $0.352$ & $0.127$ 
    \\
    & LASO~\cite{li2024laso} & \underline{$8.0$} & $69.2$ & \underline{$0.386$} & \underline{$0.118$} \\
    & \cellcolor{cell}\textbf{GEAL (Ours)} & \cellcolor{cell}$\mathbf{8.7}$ & \cellcolor{cell}$\mathbf{72.5}$ & \cellcolor{cell}$\mathbf{0.390}$ & \cellcolor{cell}$\mathbf{0.102}$ \\ 
    \bottomrule
    \end{tabular}
    }
\label{tab:piad_table}
\end{table}
\begin{table}[t]
    \centering
    \caption{The overall results of all comparative methods on the \textbf{LASO} dataset~\cite{li2024laso}. \textbf{Seen} and \textbf{Unseen} are two partitions of the dataset.  Results marked with * denote our reproduced results, following the data split reported in LASO~\cite{li2024laso}. AUC and aIOU are shown in percentage. The \textbf{best} and \underline{2nd best} scores from each metric are highlighted in \textbf{bold} and \underline{underlined}, respectively.}
    \vspace{-0.2cm}
    \resizebox{\linewidth}{!}{
    \begin{tabular}{c|c|cccc}
    \toprule
    \textbf{Type} & \textbf{Method} & \textbf{aIoU} $\uparrow$ & \textbf{AUC} $\uparrow$ & \textbf{SIM} $\uparrow$ & \textbf{MAE} $\downarrow$ 
    \\
    \midrule\midrule
    \multirow{7}{*}{\textbf{Seen}} 
    & ReferTrans~\cite{li2021referring} & $13.7$ & $79.8$ & $0.497$ & $0.124$ 
    \\
    & ReLA~\cite{liu2023gres} & $15.2$ & $78.9$ & $0.532$ & $0.118$ 
    \\
    & 3D-SPS~\cite{luo20223d} & $11.4$ & $76.2$ & $0.433$ & $0.138$ 
    \\
    & IAGNet~\cite{yang2023grounding} & $17.8$ & $82.3$ & $0.561$ & $0.109$
    \\
    & LASO~\cite{li2024laso} & \underline{$20.8$} & $\mathbf{87.3}$ & \underline{$0.629$} & \underline{$0.093$} 
    \\
    & LASO*~\cite{li2024laso} & $19.7$ & $85.2$ & $0.600$ & $0.097$ 
    \\
    & \cellcolor{cell}\textbf{GEAL (Ours)} & \cellcolor{cell}$\mathbf{22.0}$ & \cellcolor{cell}\underline{$86.7$} & \cellcolor{cell}$\mathbf{0.634}$ & \cellcolor{cell}$\mathbf{0.092}$ 
    \\ 
    \midrule
    \multirow{7}{*}{\textbf{Unseen}} 
    & ReferTrans~\cite{li2021referring} & $10.2$ & $69.1$ & $0.432$ & $0.145$ 
    \\
    & ReLA~\cite{liu2023gres} & $10.7$ & $69.7$ & $0.429$ & $0.144$ 
    \\
    & 3D-SPS~\cite{luo20223d} & $7.9$ & $68.8$ & $0.402$ & $0.158$ 
    \\
    & IAGNet~\cite{yang2023grounding} & $12.9$ & {$77.8$} & $0.443$ & {$0.129$} 
    \\
    & LASO~\cite{li2024laso} & ${14.6}$ & ${80.2}$ & ${0.507}$ & ${0.119}$ 
    \\
    & LASO*~\cite{li2024laso} & \underline{$15.6$} & \underline{$79.9$} & \underline{$0.549$} & \underline{$0.108$} 
    \\
    & \cellcolor{cell}\textbf{GEAL (Ours)} & \cellcolor{cell}$\mathbf{16.7}$ & \cellcolor{cell}$\mathbf{80.9}$ & \cellcolor{cell}$\mathbf{0.567}$ & \cellcolor{cell}$\mathbf{0.106}$ 
    \\ 
    \bottomrule
    \end{tabular}}
\label{tab:laso_table}
\end{table}

\subsection{Experimental Settings}
\noindent\textbf{Implementation Details.} Our model is implemented using PyTorch \cite{paszke2019pytorch} and is trained using the Adam optimizer \cite{kingma2015adam} with an initial learning rate of $1 \times 10^{-4}$ for $50$ epochs on a single NVIDIA A5000 GPU (with $24$ GB memory) with a batch size of $12$. A step learning rate scheduler aids convergence. Additionally, the 2D backbone DINOV2 \cite{oquab2023dinov2} remains frozen during training, while the language model RoBERTa \cite{liu2019roberta} has been fine-tuned.

\vspace{2mm}
\noindent\textbf{Datasets.} We conduct experiments on \textbf{LASO} \cite{li2024laso} and \textbf{PIAD} \cite{yang2023grounding}, both providing paired affordance and point cloud data. \textbf{LASO} contains $19,751$ point cloud-question pairs over $8,434$ objects ($23$ classes, $17$ affordance categories) with \textit{Seen} and \textit{Unseen} splits to test generalization to novel affordance-object pairs. \textbf{PIAD} comprises $7,012$ point clouds of the same categories, but some objects are entirely unseen during training, challenging the model's generalization to novel objects. Since PIAD lacks language annotations, we reuse LASO by randomly assigning questions to each affordance-object pair.

\vspace{2mm}
\noindent\textbf{Metrics.} We use four metrics to assess performance: \textbf{AUC} \cite{lobo2008auc} measures the ability to rank points correctly; \textbf{aIoU} \cite{rahman2016optimizing} quantifies the overlap between predictions and ground truth; \textbf{SIM} \cite{swain1991color} assesses the similarity by summing minimum values at each point; and \textbf{MAE} \cite{willmott2005advantages} calculates the average absolute difference between predictions and ground truth.

\vspace{2mm}
\noindent\textbf{Baselines.} We primarily compare our method with state-of-the-art approaches LASO \cite{li2024laso} and IAGNet \cite{yang2023grounding}. On \textbf{PIAD}, we evaluate against IAGNet and several image-point cloud cross-modal baselines, retraining LASO for comparison. On \textbf{LASO}, we compare with the original LASO method and other methods utilizing vision-language models for cross-modal alignment. To adapt IAGNet to LASO, its image backbone is replaced with a language model \cite{li2024laso}, keeping the rest of the architecture intact. Since the \textit{Unseen} data split of LASO is not publicly available, we reproduce it based on 
{the} descriptions {in their paper} and report our results accordingly. Further experimental details are provided in the supplementary material.

\begin{table}[t]
    \centering
    \caption{Comparison of different methods under various corruption settings on the proposed \textbf{PIAD-C} benchmark, evaluated on the \textbf{Seen} partition. \textbf{Drop-L} denotes local drop, and \textbf{Drop-G} denotes global drop; similarly, \textbf{Add-L} and \textbf{Add-G} refer to local and global addition, respectively. AUC and aIOU are reported as percentages. For each metric, the \textbf{best} scores are highlighted in \textbf{bold}.}
    \vspace{-0.2cm}
    \resizebox{\linewidth}{!}{
    \begin{tabular}{l|cc|cc|cc|cc}
    \toprule
    \multirow{2}{*}{\textbf{Type}} & \multicolumn{2}{c|}{\textbf{aIOU} $\uparrow$} & \multicolumn{2}{c|}{\textbf{AUC} $\uparrow$} & \multicolumn{2}{c|}{\textbf{SIM} $\uparrow$} & \multicolumn{2}{c}{\textbf{MAE} $\downarrow$} \\
    & LASO & \cellcolor{cell}\textbf{GEAL}
    & LASO & \cellcolor{cell}\textbf{GEAL}
    & LASO & \cellcolor{cell}\textbf{GEAL}
    & LASO & \cellcolor{cell}\textbf{GEAL} 
    \\
    \midrule\midrule
    \textbf{Scale}   & $17.6$ & \cellcolor{cell}$\mathbf{19.7}$ & $82.1$ & \cellcolor{cell}$\mathbf{82.5}$ & $0.554$ & \cellcolor{cell}$\mathbf{0.562}$ & $0.100$ & \cellcolor{cell}$\mathbf{0.097}$ 
    \\
    \textbf{Jitter}  & $14.7$ & \cellcolor{cell}$\mathbf{17.0}$ & $80.3$ & \cellcolor{cell}$\mathbf{80.6}$ & $0.501$ & \cellcolor{cell}$\mathbf{0.505}$ & $0.103$ & \cellcolor{cell}$\mathbf{0.099}$ 
    \\
    \textbf{Rotate}  & $16.7$ & \cellcolor{cell}$\mathbf{19.0}$ & $82.2$ & \cellcolor{cell}$\mathbf{82.4}$ & $0.542$ & \cellcolor{cell}$\mathbf{0.550}$ & $0.101$ & \cellcolor{cell}$\mathbf{0.097}$ 
    \\
    \textbf{Drop-L} & $10.6$ & \cellcolor{cell}$\mathbf{12.4}$ & $77.0$ & \cellcolor{cell}$\mathbf{77.2}$ & $0.470$ & \cellcolor{cell}$\mathbf{0.474}$ & $0.112$ & \cellcolor{cell}$\mathbf{0.111}$ 
    \\
    \textbf{Drop-G} & $18.7$ & \cellcolor{cell}$\mathbf{21.1}$ & $83.1$ & \cellcolor{cell}$\mathbf{83.7}$ & $0.545$ & \cellcolor{cell}$\mathbf{0.559}$ & $0.097$ & \cellcolor{cell}$\mathbf{0.094}$ 
    \\
    \textbf{Add-L}  & $15.7$ & \cellcolor{cell}$\mathbf{18.5}$ & $81.0$ & \cellcolor{cell}$\mathbf{81.1}$ & $0.525$ & \cellcolor{cell}$\mathbf{0.536}$ & $0.100$ & \cellcolor{cell}$\mathbf{0.095}$ 
    \\
    \textbf{Add-G}  & $13.4$ & \cellcolor{cell}$\mathbf{16.1}$ & $76.9$ & \cellcolor{cell}$\mathbf{77.4}$ & $0.506$ & \cellcolor{cell}$\mathbf{0.513}$ & $0.101$ & \cellcolor{cell}$\mathbf{0.098}$ 
    \\
    \bottomrule
    \end{tabular}
    }
\label{tab:robustness_piad}
\end{table}

\begin{table}[t]
    \centering
    \caption{Comparison of different methods under various corruption settings on the proposed \textbf{LASO-C} benchmark, evaluated on the \textbf{Seen} partition. \textbf{Drop-L} denotes local drop, and \textbf{Drop-G} denotes global drop; similarly, \textbf{Add-L} and \textbf{Add-G} refer to local and global addition, respectively. AUC and aIOU are reported as percentages. For each metric, the \textbf{best} scores are highlighted in \textbf{bold}.}
    \vspace{-0.2cm}
    \resizebox{\linewidth}{!}{
    \begin{tabular}{l|cc|cc|cc|cc}
    \toprule
    \multirow{2}{*}{\textbf{Type}} & \multicolumn{2}{c|}{\textbf{aIOU} $\uparrow$} & \multicolumn{2}{c|}{\textbf{AUC} $\uparrow$} & \multicolumn{2}{c|}{\textbf{SIM} $\uparrow$} & \multicolumn{2}{c}{\textbf{MAE} $\downarrow$} 
    \\
    & LASO & \cellcolor{cell}\textbf{GEAL}
    & LASO & \cellcolor{cell}\textbf{GEAL}
    & LASO & \cellcolor{cell}\textbf{GEAL}
    & LASO & \cellcolor{cell}\textbf{GEAL}
    \\
    \midrule\midrule
    \textbf{Scale}   & $18.7$ & \cellcolor{cell}$\mathbf{21.0}$ & $84.6$ & \cellcolor{cell}$\mathbf{85.3}$ & $0.590$ & \cellcolor{cell}$\mathbf{0.600}$ & $0.103$ & \cellcolor{cell}$\mathbf{0.100}$ 
    \\
    \textbf{Jitter}  & $15.4$ & \cellcolor{cell}$\mathbf{17.8}$ & $81.3$ & \cellcolor{cell}$\mathbf{81.9}$ & $0.516$ & \cellcolor{cell}$\mathbf{0.517}$ & $0.107$ & \cellcolor{cell}$\mathbf{0.106}$ 
    \\
    \textbf{Rotate}  & $17.8$ & \cellcolor{cell}$\mathbf{19.8}$ & $83.6$ & \cellcolor{cell}$\mathbf{84.3}$ & $0.572$ & \cellcolor{cell}$\mathbf{0.573}$ & $0.101$ & \cellcolor{cell}$\mathbf{0.100}$ 
    \\
    \textbf{Drop-L} & $12.6$ & \cellcolor{cell}$\mathbf{13.3}$ & $79.3$ & \cellcolor{cell}$\mathbf{80.0}$ & $0.466$ & \cellcolor{cell}$\mathbf{0.484}$ & $0.122$ & \cellcolor{cell}$\mathbf{0.110}$ 
    \\
    \textbf{Drop-G} & $18.4$ & \cellcolor{cell}$\mathbf{20.9}$ & $83.5$ & \cellcolor{cell}$\mathbf{85.2}$ & $0.565$ & \cellcolor{cell}$\mathbf{0.567}$ & $0.099$ & \cellcolor{cell}$\mathbf{0.095}$ 
    \\
    \textbf{Add-L}  & $17.6$ & \cellcolor{cell}$\mathbf{20.2}$ & $82.7$ & \cellcolor{cell}$\mathbf{84.4}$ & $0.566$ & \cellcolor{cell}$\mathbf{0.572}$ & $0.103$ & \cellcolor{cell}$\mathbf{0.100}$ 
    \\
    \textbf{Add-G}  & $16.7$ & \cellcolor{cell}$\mathbf{19.0}$ & $81.1$ & \cellcolor{cell}$\mathbf{83.4}$ & $0.549$ & \cellcolor{cell}$\mathbf{0.575}$ & $0.108$ & \cellcolor{cell}$\mathbf{0.097}$ \\
    \bottomrule
    \end{tabular}
    }
\label{tab:robustness_laso}
\end{table}
\begin{table}[t]
    \centering
    \caption{Ablation study on the impact of different components in \textbf{GEAL} on the \textbf{PIAD} dataset~\cite{yang2023grounding}. \textbf{Seen} and \textbf{Unseen} are two partitions of the dataset. \textbf{2D} denotes the use of the 2D baseline with a weighted sum mapping back to 3D. \textbf{3D} represents the 3D baseline. \textbf{CAM} is the consistency alignment module. \textbf{GAFM} is the granularity-adaptive fusion module. AUC and aIoU are shown in percentage. The \textbf{best} and \underline{second best} scores from each metric are highlighted in \textbf{bold} and \underline{underlined}, respectively.}
    \vspace{-0.2cm}
    \resizebox{\linewidth}{!}{
    \begin{tabular}{c|c|c|c|c|cccc}
    \toprule
    \textbf{Type} & \textbf{2D} & \textbf{3D} & \textbf{CAM} &\textbf{GAFM} & \textbf{aIoU} $\uparrow$ & \textbf{AUC} $\uparrow$ & \textbf{SIM} $\uparrow$ & \textbf{MAE} $\downarrow$ 
    \\
    \midrule\midrule
    \multirow{4}{*}{\textbf{Seen}} 
    & \cmark & \xmark & \xmark & \xmark & $19.2$ & $80.5$ & $0.567$ & $0.101$ \\
    & \xmark & \cmark & \xmark & \xmark & $19.5$ & $83.5$ & $0.585$ & $0.097$ \\
    & \cmark & \cmark & \cmark & \xmark & \underline{$22.0$} & \underline{$84.4$} & \underline{$0.592$} & \underline{$0.094$} \\
    & \cellcolor{cell}\cmark & \cellcolor{cell}\cmark & \cellcolor{cell}\cmark & \cellcolor{cell}\cmark & \cellcolor{cell}$\mathbf{22.5}$ & \cellcolor{cell}$\mathbf{85.0}$ & \cellcolor{cell}$\mathbf{0.600}$ & \cellcolor{cell}$\mathbf{0.092}$ \\
    \midrule
    \multirow{4}{*}{\textbf{Unseen}} 
    & \cmark & \xmark & \xmark  & \xmark & $8.5$ & $70.8$ & $0.357$ & $0.112$ \\
    & \xmark & \cmark & \xmark  & \xmark & $8.0$ & $69.2$ & \underline{$0.386$} & $0.118$ \\
    & \cmark & \cmark & \cmark  & \xmark & \underline{$8.6$} & \underline{$71.2$} & $0.371$ & \underline{$0.105$} \\
    & \cellcolor{cell}\cmark & \cellcolor{cell}\cmark & \cellcolor{cell}\cmark  & \cellcolor{cell}\cmark & \cellcolor{cell}$\mathbf{8.7}$ & \cellcolor{cell}$\mathbf{72.5}$ & \cellcolor{cell}$\mathbf{0.390}$ & \cellcolor{cell}$\mathbf{0.102}$ \\
    \bottomrule
    \end{tabular}}
\label{tab:ablation_component}
\end{table}

\begin{table}[t]
    \centering
    \caption{Ablation study on the configuration of Gaussian Splatting parameters in \textbf{GEAL} on the \textbf{PIAD} dataset~\cite{yang2023grounding}. ${r}$ denotes the resolution, ${V}$ is the number of views, and \textbf{prompt} indicates whether a view-dependent prompt is used. \textbf{Seen} and \textbf{Unseen} are two partitions of the dataset. AUC and aIoU are shown in percentage. The \textbf{best} and \underline{2nd best} scores from each metric are highlighted in \textbf{bold} and \underline{underlined}, respectively.}
    \vspace{-0.2cm}
    \resizebox{\linewidth}{!}{
    \begin{tabular}{c|c|c|c|cccc}
    \toprule
    \textbf{Type} & ${r}$ & ${V}$ & \textbf{prompt} & \textbf{aIoU} $\uparrow$ & \textbf{AUC} $\uparrow$ & \textbf{SIM} $\uparrow$ & \textbf{MAE} $\downarrow$ 
    \\
    \midrule\midrule
    \multirow{5}{*}{\textbf{Seen}} 
    & $112$ & $6$ & \xmark & $20.2$ & $83.5$ & $0.566$ & $0.112$ \\
    & $112$ & $12$ & \xmark & $21.4$ & $83.8$ & $0.578$ & $0.105$ \\
    & \cellcolor{cell}$112$ & \cellcolor{cell}$12$ & \cellcolor{cell}\cmark & \cellcolor{cell}$22.5$ & \cellcolor{cell}$85.0$ & \cellcolor{cell}$0.600$ & \cellcolor{cell}$0.092$  \\
    & $112$ & $14$ & \cmark & $22.5$ & $85.2$ & $0.599$ & $0.092$ \\
    & $224$ & $14$ & \cmark & $\mathbf{22.9}$ & $\mathbf{86.1}$ & $\mathbf{0.603}$ & $\mathbf{0.089}$ \\
    \midrule
    \multirow{5}{*}{\textbf{Unseen}} 
    & $112$ & $6$ & \xmark & $7.0$ & $70.7$ & $0.355$ & $0.108$ \\
    & $112$ & $12$ & \xmark & $7.5$ & $71.9$ & $0.390$ & $0.106$ \\
    & \cellcolor{cell}$112$ & \cellcolor{cell}$12$ & \cellcolor{cell}\cmark & \cellcolor{cell}$8.7$ & \cellcolor{cell}$72.5$ & \cellcolor{cell}$0.390$ & \cellcolor{cell}$0.102$ \\
    & $112$ & $14$ & \cmark & $8.9$ & $72.8$ & $0.391$ & $0.098$ \\
    & $224$ & $14$ & \cmark & $\mathbf{9.2}$ & $\mathbf{73.0}$ & $\mathbf{0.394}$ & $\mathbf{0.095}$ \\
    \bottomrule
    \end{tabular}}
\label{tab:ablation_gaussian}
\end{table}
\begin{figure}[t]
    \centering
    \includegraphics[width=0.98\linewidth]{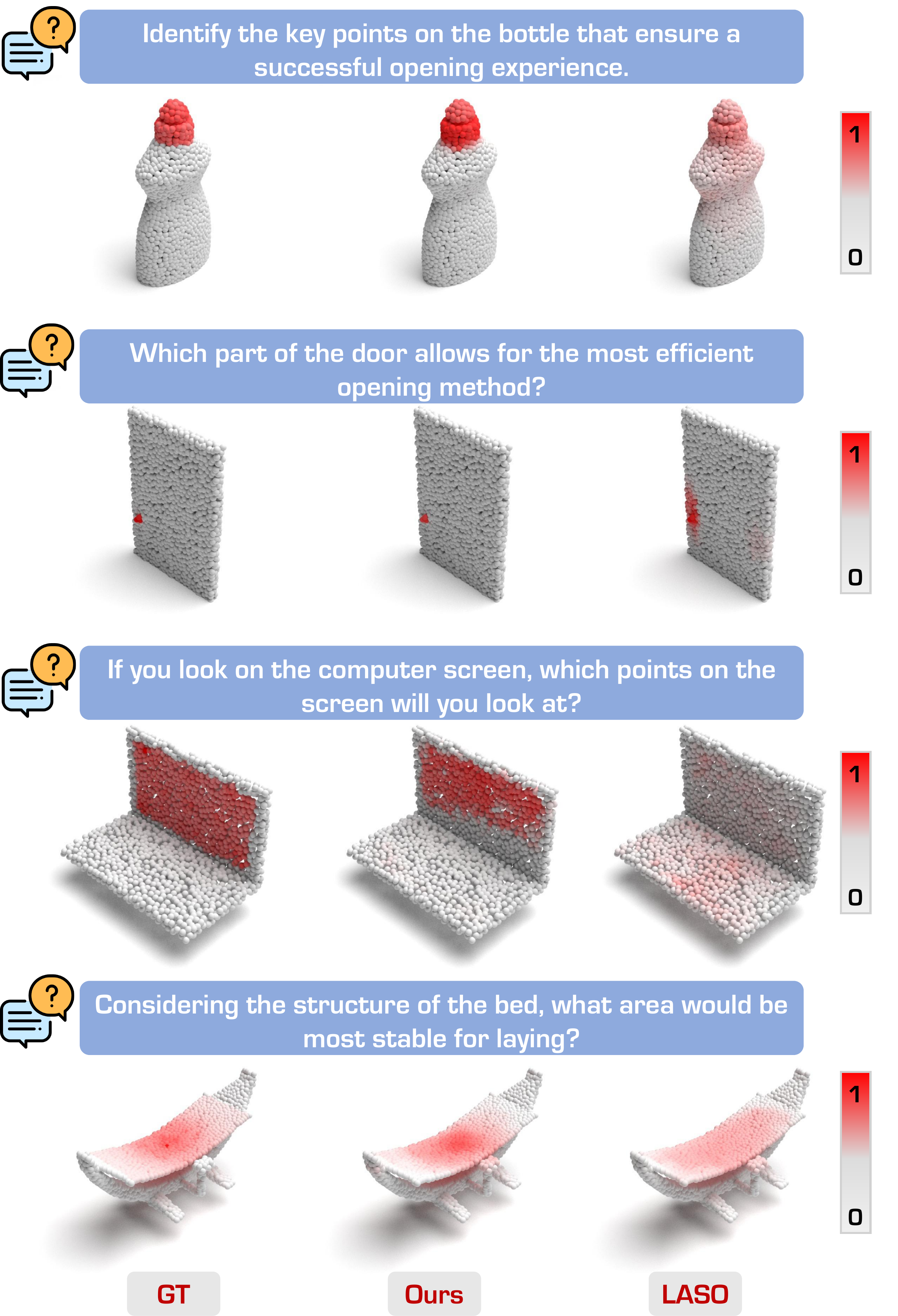}
    \caption{
     Qualitative comparisons between \textbf{GEAL} and LASO \cite{li2024laso} on the PIAD \cite{yang2023grounding} dataset. 
     Top two rows display results on 
     \textit{seen} partition, while 
     bottom two rows show results on 
     \textit{unseen} partition. Our method demonstrates strong generalization on both seen and unseen partitions. 
     {\cf supplementary material} for more examples.
    }
  \label{fig:seen_unseen}
\end{figure}

\subsection{Comparisons to State-of-the-Art Methods}

\noindent\textbf{Seen Categories:} In \cref{tab:piad_table} and \cref{tab:laso_table}, we present the performance of our 
\textbf{GEAL} compared to state-of-the-art approaches on the PIAD and LASO datasets under the \textit{Seen} category setting. On the PIAD dataset, GEAL achieves the highest scores across all evaluation metrics, surpassing the previous best method, IAGNet \cite{yang2023grounding}. Similarly, on the LASO dataset, GEAL outperforms LASO \cite{li2024laso} on the majority of metrics. These results highlight the effectiveness of GEAL in leveraging the rich semantic understanding from pre-trained 2D models through Gaussian splatting. The granularity-adaptive fusion and 2D-3D consistency alignment modules enable multi-granularity feature fusion and efficient knowledge transfer between the 2D and 3D modalities, enhancing the 
ability {of the model} to accurately predict affordance regions on the seen categories.

\vspace{2mm}
\noindent\textbf{Unseen Categories:} The \textit{Unseen} category setting evaluates the 
{generalization ability of the model} to novel objects not encountered during training. On both the PIAD (\cref{tab:piad_table}) and LASO (\cref{tab:laso_table}) datasets, \textbf{GEAL}  consistently outperforms all baselines across metrics. Although the absolute performance values are lower due to the increased difficulty of unseen categories, GEAL maintains a performance edge over the baselines. This demonstrates that GEAL effectively generalizes to unseen categories, a result attributed to the integration of the 2D branch with a pretrained foundation model backbone and the cross-modal consistency alignment between the 2D and 3D branches. Qualitative comparisons with LASO on PIAD are shown in ~\cref{fig:seen_unseen}.

\vspace{2mm}
\noindent\textbf{Robustness on Corrupt Data:} To assess robustness under real-world conditions, we compare \textbf{GEAL} with LASO on the proposed PIAD-C and LASO-C benchmarks after training on clean data. As shown in \cref{tab:robustness_piad} and ~\cref{tab:robustness_laso}, GEAL consistently outperforms LASO across all corruption types and evaluation metrics. GEAL demonstrates superior resilience under various corruptions, achieving higher AUC and SIM scores while maintaining lower MAE values. This consistent outperformance indicates that the architecture {of GEAL} effectively mitigates the impact of data degradation. The robustness improvements are attributed to our dual-branch architecture and the 2D-3D consistency alignment module. By leveraging the robustness of pre-trained 2D models and enforcing cross-modal consistency, GEAL maintains high performance even when faced with corrupted or noisy 3D data.

\subsection{Ablation Study}
\noindent\textbf{Component Analysis.} 
As shown in Table~\ref{tab:ablation_component}, we conduct an ablation study on the PIAD dataset \cite{yang2023grounding} to evaluate the effectiveness of each component in our proposed \textbf{GEAL} framework. We examine the impact of using only the 2D baseline with a weighted sum mapping back to 3D using the inverse process of \cref{eq:rendering_3dgs} (\ie, \textbf{2D}), only the 3D baseline (\ie, \textbf{3D}), the consistency alignment module (\textbf{CAM}), and the granularity-adaptive fusion module (\textbf{GAFM}). Both the 2D and 3D baselines use only the last-layer features from their respective visual backbones to fuse with textual features without considering granularity. The results show that using only the 2D branch or only the 3D branch yields similar baseline performance. Integrating both branches with the consistency alignment module (\textbf{CAM}) leads to a noticeable improvement. Finally, our full model incorporating all components, including the granularity-adaptive fusion module (\textbf{GAFM}), achieves the best performance on both the \textit{Seen} and \textit{Unseen} sets. This demonstrates the effectiveness of our dual-branch architecture and underscores the importance of granularity-adaptive fusion and cross-modal consistency in enhancing the
generalization capability of the model.

\vspace{2mm}
\noindent\textbf{Gaussian Splatting Configuration Tuning.}
{As shown in Table~\ref{tab:ablation_gaussian},} we further investigate the impact of different Gaussian splatting configurations on the performance of our model. Specifically, we vary the rendering resolution ($r$), the number of views ($V$), and the inclusion of view-dependent prompts. The results indicate that increasing the number of views from $6$ to $12$ slightly improves performance, suggesting that additional viewpoints provide richer information for affordance learning. Incorporating view-dependent prompts significantly boosts performance, particularly on the \textit{Seen} set, highlighting the importance of semantic guidance in our framework. Increasing the resolution from $112$ to $224$ yields only marginal gains, indicating that our model is robust to resolution changes and that higher resolutions offer diminishing returns.  Balancing effectiveness and efficiency, we opt for a configuration of $r=112$, $V=12$, and the use of view-dependent prompts.
\section{Conclusion}
\label{sec:conclusion}
In conclusion, we present \textbf{GEAL}, a framework that improves the generalization and robustness of 3D affordance learning by leveraging large-scale pre-trained 2D models. Through a dual-branch architecture with Gaussian splatting, GEAL maps 3D point clouds to 2D representations, enabling realistic renderings from sparse data. The granularity-adaptive fusion and 2D-3D consistency alignment modules support cross-modal alignment and knowledge transfer, allowing the 3D branch to leverage rich semantic information from pre-trained 2D models. Experiments on public datasets and our corruption-based benchmarks show that GEAL consistently outperforms existing methods, demonstrating robust affordance prediction under varied conditions.

\appendix
\renewcommand{\appendixname}{Appendix~\Alph{section}}

\section*{Appendix}
\startcontents[appendices]
\printcontents[appendices]{l}{1}{\setcounter{tocdepth}{3}}
\vspace{0.5cm}

\section{Corrupt Data Benchmark}
\label{sec:corrupt_data_benchmark}
The robustness of models under real-world corruptions is a critical challenge in 3D point cloud analysis and 3D affordance learning~\cite{hendrycks2019benchmarking,ren2022benchmarking, wang2023sample,kim2021point}. Unlike other 3D representations, point clouds often face various distortions caused by sensor inaccuracies, environmental complexities, and post-processing artifacts, which significantly impact downstream tasks~\cite{lee2021regularization, wang2024target, kong2023robo3d}. For 3D affordance learning, ensuring robustness is paramount, as affordances are highly sensitive to object geometry and spatial details.

\subsection{Corruption \& Severity Level Settings}
To standardize evaluation, we introduce a taxonomy of \textbf{seven atomic corruption types} -- \textit{Scale}, \textit{Jitter}, \textit{Rotate}, \textit{Drop Global}, \textit{Drop Local}, \textit{Add Global}, \textit{Add Local} -- each simulating distinct real-world perturbations. These atomic corruptions simplify complex scenarios into controllable factors, enabling systematic analysis across \textbf{five levels of severity}. By providing a unified framework for benchmarking, we facilitate consistent and comprehensive assessment of model robustness, setting the stage for more resilient 3D affordance learning methods.

\begin{table}[t]
    \centering
    \caption{Detailed statistics of the proposed \textbf{PIAD-C} dataset, showing the object categories, their corresponding affordance types, and the number of object-affordance pairings for each category.}
    \vspace{-0.2cm}
    \resizebox{\linewidth}{!}{
    \begin{tabular}{c|r|c|c}
    \toprule
    \textbf{\#} & \textbf{Object Category} & \textbf{Affordance Type} & \textbf{Data} \\ 
    \midrule\midrule
    $1$ & \textbf{Earphone} \textcolor{BurntOrange}{$\bullet$} & listen, grasp & $70$ \\ 
    \rowcolor{gray!10}$2$ & \textbf{Bag} \textcolor{BurntOrange}{$\bullet$} & contain, open, grasp, lift & $50$ \\ 
    $3$ & \textbf{Chair} \textcolor{BurntOrange}{$\bullet$} & move, support, sit & $587$ \\ 
    \rowcolor{gray!10}$4$ & \textbf{Refrigerator} \textcolor{BurntOrange}{$\bullet$} & contain, open & $53$ \\ 
    $5$ & \textbf{Knife} \textcolor{BurntOrange}{$\bullet$} & stab, cut, grasp & $138$ \\ 
    \rowcolor{gray!10}$6$ & \textbf{Dishwasher} \textcolor{BurntOrange}{$\bullet$} & contain, open & $39$ \\ 
    $7$ & \textbf{Keyboard} \textcolor{BurntOrange}{$\bullet$} & press & $25$ \\ 
    \rowcolor{gray!10}$8$ & \textbf{Scissors} \textcolor{BurntOrange}{$\bullet$} & stab, cut, grasp & $29$ \\ 
    $9$ & \textbf{Table} \textcolor{BurntOrange}{$\bullet$} & move, support & $194$ \\ 
    \rowcolor{gray!10}$10$ & \textbf{StorageFurniture} \textcolor{BurntOrange}{$\bullet$} & contain, open & $92$ \\ 
    $11$ & \textbf{Bottle} \textcolor{BurntOrange}{$\bullet$} & contain, wrap\_grasp, open, grasp, pour & $273$ \\ 
    \rowcolor{gray!10}$12$ & \textbf{Bowl} \textcolor{BurntOrange}{$\bullet$} & contain, wrap-grasp, pour & $83$ \\ 
    $13$ & \textbf{Microwave} \textcolor{BurntOrange}{$\bullet$} & contain, open & $47$ \\ 
    \rowcolor{gray!10}$14$ & \textbf{Display} \textcolor{BurntOrange}{$\bullet$} & display & $52$ \\ 
    $15$ & \textbf{TrashCan} \textcolor{BurntOrange}{$\bullet$} & contain, open, pour & $69$ \\ 
    \rowcolor{gray!10}$16$ & \textbf{Hat} \textcolor{BurntOrange}{$\bullet$} & wear, grasp & $66$ \\ 
    $17$ & \textbf{Clock} \textcolor{BurntOrange}{$\bullet$} & display & $9$ \\ 
    \rowcolor{gray!10}$18$ & \textbf{Door} \textcolor{BurntOrange}{$\bullet$} & open, push & $47$ \\ 
    $19$ & \textbf{Mug} \textcolor{BurntOrange}{$\bullet$} & contain, wrap\_grasp, grasp, pour & $126$ \\ 
    \rowcolor{gray!10}$20$ & \textbf{Faucet} \textcolor{BurntOrange}{$\bullet$} & open, grasp & $95$ \\ 
    $21$ & \textbf{Vase} \textcolor{BurntOrange}{$\bullet$} & contain, wrap-grasp, pour & $134$ \\ 
    \rowcolor{gray!10}$22$ & \textbf{Laptop} \textcolor{BurntOrange}{$\bullet$} & press, display & $112$ \\ 
    $23$ & \textbf{Bed} \textcolor{BurntOrange}{$\bullet$} & lay, support, sit & $84$ \\ 
    \midrule
    \rowcolor{gray!10}\textbf{Total} & $\mathbf{23}$ \textbf{Categories} & $\mathbf{17}$ \textbf{Affordance Types} & $2474$ \\
    \bottomrule
    \end{tabular}
    }
    \label{tab:piad_c_stat}
\end{table}

\begin{table}[t]
    \centering
    \caption{Detailed statistics of the proposed \textbf{LASO-C} dataset, showing the object categories, their corresponding affordance types, and the number of distinct objects for each category.}
    \vspace{-0.2cm}
    \resizebox{\linewidth}{!}{
    \begin{tabular}{c|r|c|c}
    \toprule
    \textbf{\#} & \textbf{Object Category} & \textbf{Affordance} & \textbf{Data} \\ 
    \midrule\midrule
    $1$ & \textbf{Door} \textcolor{Emerald}{$\bullet$} & open, push, pull & $35$ \\ 
    \rowcolor{gray!10}$2$ & \textbf{Clock} \textcolor{Emerald}{$\bullet$} & display & $34$ \\ 
    $3$ & \textbf{Dishwasher} \textcolor{Emerald}{$\bullet$} & open, contain & $20$ \\ 
    \rowcolor{gray!10}$4$ & \textbf{Earphone} \textcolor{Emerald}{$\bullet$} & listen, grasp & $28$ \\ 
    $5$ & \textbf{Vase} \textcolor{Emerald}{$\bullet$} & contain, pour, wrap-grasp & $167$ \\ 
    \rowcolor{gray!10}$6$ & \textbf{Knife} \textcolor{Emerald}{$\bullet$} & stab, grasp, cut & $59$ \\ 
    $7$ & \textbf{Bowl} \textcolor{Emerald}{$\bullet$} & contain, pour, wrap\_grasp & $36$ \\ 
    \rowcolor{gray!10}$8$ & \textbf{Bag} \textcolor{Emerald}{$\bullet$} & open, contain, lift, grasp & $25$ \\ 
    $9$ & \textbf{Faucet} \textcolor{Emerald}{$\bullet$} & open, grasp & $80$ \\ 
    \rowcolor{gray!10}$10$ & \textbf{Scissors} \textcolor{Emerald}{$\bullet$} & stab, grasp, cut & $11$ \\ 
    $11$ & \textbf{Display} \textcolor{Emerald}{$\bullet$} & display & $58$ \\ 
    \rowcolor{gray!10}$12$ & \textbf{Chair} \textcolor{Emerald}{$\bullet$} & sit, support, move & $858$ \\ 
    $13$ & \textbf{Bottle} \textcolor{Emerald}{$\bullet$} & grasp, wrap\_grasp, open, contain, pour & $122$ \\ 
    \rowcolor{gray!10}$14$ & \textbf{Microwave} \textcolor{Emerald}{$\bullet$} & open, contain & $23$ \\ 
    $15$ & \textbf{StorageFurniture} \textcolor{Emerald}{$\bullet$} & open, contain & $183$ \\ 
    \rowcolor{gray!10}$16$ & \textbf{Refrigerator} \textcolor{Emerald}{$\bullet$} & open, contain & $23$ \\ 
    $17$ & \textbf{Mug} \textcolor{Emerald}{$\bullet$} & contain, grasp, pour, wrap-grasp & $45$ \\ 
    \rowcolor{gray!10}$18$ & \textbf{Keyboard} \textcolor{Emerald}{$\bullet$} & press & $10$ \\ 
    $19$ & \textbf{Table} \textcolor{Emerald}{$\bullet$} & support, move & $431$ \\ 
    \rowcolor{gray!10}$20$ & \textbf{Bed} \textcolor{Emerald}{$\bullet$} & sit, support, lay & $36$ \\ 
    $21$ & \textbf{Hat} \textcolor{Emerald}{$\bullet$} & wear, grasp & $26$ \\ 
    \rowcolor{gray!10}$22$ & \textbf{Laptop} \textcolor{Emerald}{$\bullet$} & display, press & $55$ \\ 
    $23$ & \textbf{TrashCan} \textcolor{Emerald}{$\bullet$} & open, contain, pour & $51$ \\ 
    \midrule
    \rowcolor{gray!10}\textbf{Total} & $\mathbf{23}$ \textbf{Categories} & $\mathbf{17}$ \textbf{Affordance Types} & $2416$ \\ 
    \bottomrule
    \end{tabular}
    }
    \label{tab:laso_c_stat}
\end{table}

Below, we detail the construction methodology for each corruption type:
\begin{itemize}
    \item \textbf{Jitter}
    \begin{itemize}
        \item \textit{Description}: Adds Gaussian noise to perturb each point's X, Y, and Z coordinates.
        \item \textit{Mathematical Formulation}: For each point, a noise $\epsilon \sim \mathcal{N}(0, \sigma^2)$ is added independently to X, Y, and Z.
        \item \textit{Severity Levels}: The standard deviation $\sigma$ varies as:
        \[
        \sigma \in \{0.01, 0.02, 0.03, 0.04, 0.05\}.
        \]
    \end{itemize}

    \item \textbf{Scale}
    \begin{itemize}
        \item \textit{Description}: Applies random scaling independently to the X, Y, and Z axes.
        \item \textit{Mathematical Formulation}: Each axis is scaled by a factor $s \sim \mathcal{U}\left(\frac{1}{S}, S\right)$, where $S$ determines the range of scaling.
        \item \textit{Severity Levels}: The range of $S$ is:
        \[
        S \in \{1.6, 1.7, 1.8, 1.9, 2.0\}.
        \]
        After scaling, the point cloud is re-normalized to fit within a unit sphere.
    \end{itemize}

    \item \textbf{Rotate}
    \begin{itemize}
        \item \textit{Description}: Introduces random rotation to the point cloud.
        \item \textit{Mathematical Formulation}: The rotation is specified by Euler angles $(\alpha, \beta, \gamma)$, where:
        \[
        \alpha, \beta, \gamma \sim \mathcal{U}(-\theta, \theta).
        \]
        \item \textit{Severity Levels}: The angle range $\theta$ is:
        \[
        \theta \in \{\pi/30, \pi/15, \pi/10, \pi/7.5, \pi/6\}.
        \]
        This approach does not guarantee uniform sampling in $\text{SO}(3)$, but provides sufficient variation to simulate diverse rotations.
    \end{itemize}

    \item \textbf{Drop Global}
    \begin{itemize}
        \item \textit{Description}: Randomly removes a percentage of points from the point cloud.
        \item \textit{Method}: Shuffle all points and drop the last $N \cdot \rho$ points, where $N = 2048$ is the total number of points.
        \item \textit{Severity Levels}: The proportion $\rho$ is:
        \[
        \rho \in \{0.25, 0.375, 0.5, 0.675, 0.75\}.
        \]
    \end{itemize}

    \item \textbf{Drop Local}
    \begin{itemize}
        \item \textit{Description}: Removes points in clusters around randomly selected local regions.
        \item \textit{Method}:
        \begin{enumerate}
            \item Randomly choose the number of local regions $C \sim \mathcal{U}\{1, 8\}$.
            \item For each region $i$:
            \begin{itemize}
                \item Randomly select a local center.
                \item Assign a cluster size $N_i$.
                \item Drop the $N_i$-nearest neighbor points to the center.
            \end{itemize}
            \item Repeat for $C$ regions.
        \end{enumerate}
        \item \textit{Severity Levels}: The total number of points to drop $K$ is:
        \[
        K \in \{100, 200, 300, 400, 500\}.
        \]
    \end{itemize}

    \item \textbf{Add Global}
    \begin{itemize}
        \item \textit{Description}: Uniformly samples additional points inside a unit sphere and appends them to the point cloud. The added points are treated as noise and assigned a label of $0$.
        \item \textit{Method}: Sample $K$ random points within a unit sphere.
        \item \textit{Severity Levels}: The total number of added points $K$ is:
        \[
        K \in \{10, 20, 30, 40, 50\}.
        \]
    \end{itemize}

    \item \textbf{Add Local}
    \begin{itemize}
        \item \textit{Description}: Adds clusters of points around randomly selected local regions. The added points are treated as noise and assigned a label of $0$.
        \item \textit{Method}:
        \begin{enumerate}
            \item Shuffle points and select $C \sim \mathcal{U}\{1, 8\}$ as the number of local centers.
            \item For each center $i$:
            \begin{itemize}
                \item Define a cluster size $N_i$.
                \item Generate neighboring points' coordinates from:
                \[
                \mathcal{N}(\mu_i, \sigma_i^2 I),
                \]
                where $\mu_i$ is the $i$-th local center, and $\sigma_i \sim \mathcal{U}(0.075, 0.125)$.
            \end{itemize}
            \item Append generated points to the cloud one cluster at a time.
        \end{enumerate}
        \item \textit{Severity Levels}: The total number of added points $K$ is:
        \[
        K \in \{100, 200, 300, 400, 500\}.
        \]
    \end{itemize}

\end{itemize}

\begin{figure*}[t]
    \centering
    \includegraphics[width=\linewidth]{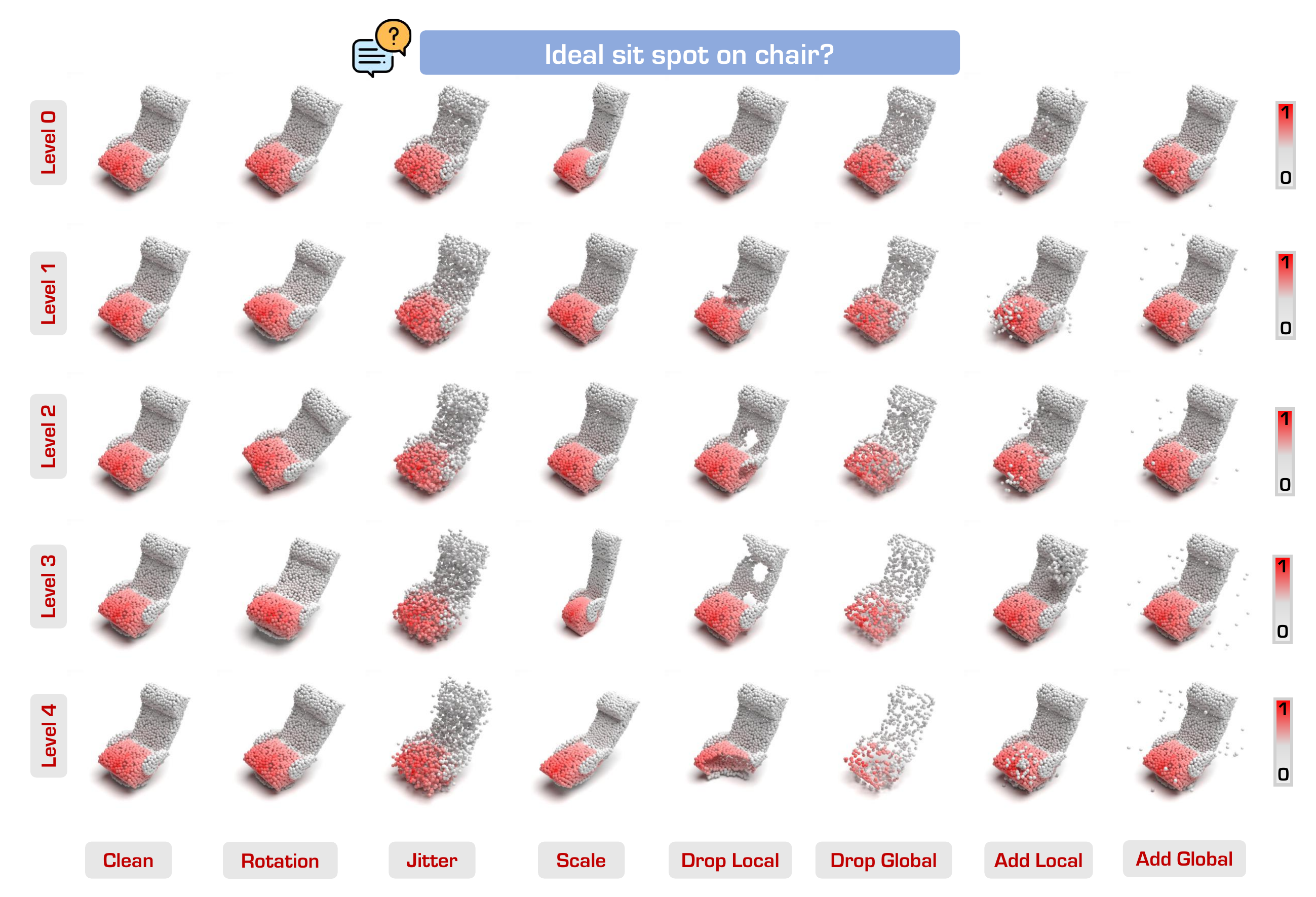}
    \caption{Visualization examples of the PIAD-C dataset. We show 7 corruption types across 5 severity levels.
    }
  \label{fig:append_benchmark_1}
\end{figure*}

\begin{figure*}[t]
    \centering
    \includegraphics[width=\linewidth]{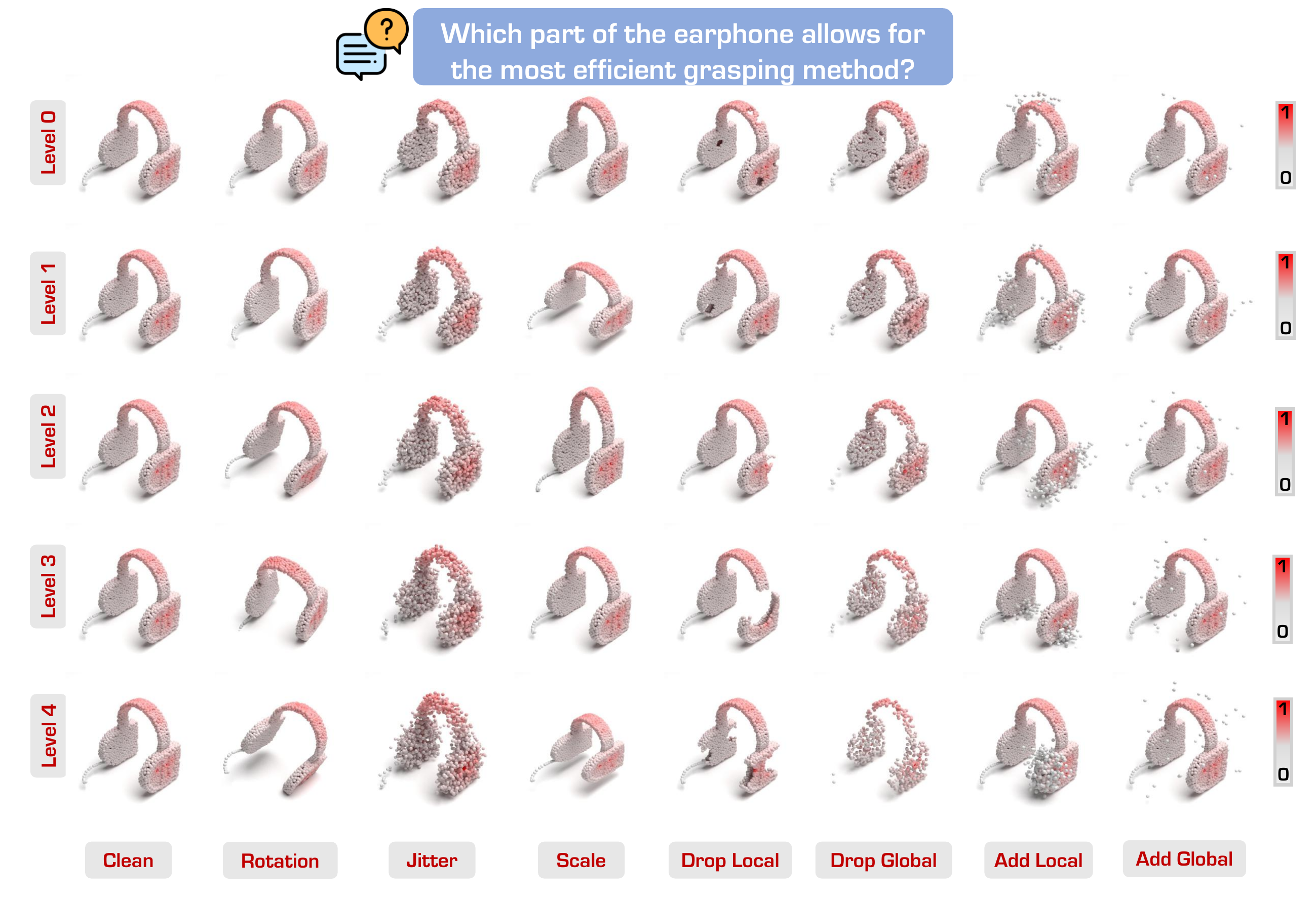}
    \caption{Visualization examples of the LASO-C dataset. We show 7 corruption types across 5 severity levels.
    }
  \label{fig:append_benchmark_2}
\end{figure*}

\subsection{The PIAD-C Dataset}
Our proposed PIAD-C dataset is constructed from the test set of the \textbf{Seen} partition in PIAD~\cite{yang2023grounding}, specifically designed to evaluate the robustness of affordance detection models under various corruption scenarios. This dataset includes a total of $2,474$ object-affordance pairings, representing $17$ affordance categories and $23$ object categories, and with $1,012$ distinct clean object shapes. Comprehensive statistics, detailing object categories, their corresponding affordance categories, and the number of object-affordance pairings, are presented in ~\cref{tab:piad_c_stat}. We include additional visualization examples for the PIAD-C dataset in ~\cref{fig:append_benchmark_1}.

\subsection{The LASO-C Dataset} 
Our proposed LASO-C dataset is derived from the test set of the \textbf{Seen} partition in LASO~\cite{li2024laso}, focusing on evaluating model robustness against point cloud corruptions. This dataset comprises $2,416$ object-affordance pairings, covering $17$ affordance categories and $23$ object categories, with a total of $1,035$ distinct clean object shapes. 

The comprehensive statistics, detailing object categories, their corresponding affordance categories, and the number of object-affordance pairings, are presented in ~\cref{tab:laso_c_stat}. We include additional visualization examples for the LASO-C dataset in ~\cref{fig:append_benchmark_2}.

\section{Benchmark Configuration}
In this section, we elaborate in more detail on the configurations and evaluations of the proposed robust 3D affordance learning benchmark.

\subsection{Datasets} 
We conduct experiments primarily on the \textbf{LASO}\citep{li2024laso} and \textbf{PIAD}\citep{yang2023grounding} datasets, both of which provide paired affordance and point cloud data for evaluating 3D affordance learning.

\vspace{2mm}
\noindent \textbf{LASO.} This dataset is a pioneering benchmark designed to enable language-guided affordance segmentation of 3D objects. It includes \textbf{$\mathbf{19,751}$ point cloud-question pairs} across \textbf{$\mathbf{8,434}$ unique object shapes}, spanning \textbf{$\mathbf{23}$ object categories} and \textbf{$\mathbf{17}$ affordance types}. Derived from \textbf{3D-AffordanceNet}~\citep{deng20213d}, the dataset pairs 3D object point clouds with questions that were carefully crafted by human experts and augmented using \textbf{GPT-4}. This process incorporates principles of \textit{contextual enrichment}, \textit{concise phrasing}, and \textit{structural diversity}, enhancing the linguistic variety and complexity of the dataset.

The LASO dataset introduces two distinct evaluation settings:
\begin{itemize}
    \item \textbf{Seen Setting:} Models are trained and tested on overlapping object-affordance combinations, ensuring that both the object classes and affordance types in the training set are also present in the test set.
    \item \textbf{Unseen Setting:} This setting is designed to evaluate generalization capabilities. Certain object-affordance combinations (\eg, ``grasp-mug'') are excluded during training but appear in testing. This setting challenges models to transfer affordance knowledge learned from seen combinations (\eg, ``grasp-bag'') to novel combinations, promoting robust generalization.
\end{itemize}

These settings promote a comprehensive evaluation of models' abilities to generalize affordance knowledge to unseen object-affordance pairings, a critical aspect for real-world deployment. The dataset also emphasizes {diverse affordance scales and shapes}, presenting significant challenges for perception models. By addressing the semantic limitations of traditional visual-only 3D affordance datasets, LASO bridges the gap between 3D perception and natural language understanding, encouraging {cross-modal learning}. This integration fosters advancements in embodied AI, enabling tasks that require nuanced reasoning and action in real-world environments.

\vspace{2mm}
\noindent \textbf{PIAD.} The Point-Image Affordance Dataset (PIAD)~\citep{yang2023grounding} is specifically curated to advance the task of grounding 3D object affordances using 2D interactions. PIAD consists of \textbf{$\mathbf{7,012}$ point clouds} and \textbf{$\mathbf{5,162}$ images}, spanning \textbf{$\mathbf{23}$ object classes} and \textbf{$\mathbf{17}$ affordance categories}. Unlike other datasets, PIAD pairs point clouds with images that demonstrate corresponding affordances. For example, a point cloud of a ``Chair'' affords ``Sit,'' and its paired image depicts a person sitting on a chair. These cross-modal pairings ensure consistency in affordance relationships while leveraging distinct modalities.

PIAD introduces two distinct evaluation settings:
\begin{itemize}
    \item \textbf{Seen Setting:} In this setting, both objects and affordances in the training and testing sets are consistent. Point clouds and images of the same object categories and affordance types are included during training, allowing models to learn affordance relationships in a supervised manner. This standard evaluation setting enables benchmarking on familiar object-affordance combinations.

    \item \textbf{Unseen Setting:} The Unseen partition presents a more challenging evaluation by excluding certain object categories from the training set entirely. For instance, some object categories are entirely unseen during training. This partition tests the ability of methods to transfer affordance knowledge across completely novel object instances and contexts, simulating real-world scenarios where interaction data is sparse or varied.
\end{itemize}

Annotations in PIAD include detailed affordance labels for point clouds, represented as heatmaps indicating the likelihood of affordance at each point. Paired images are annotated with bounding boxes for interactive subjects and objects, along with affordance category labels. This comprehensive annotation schema supports diverse affordance-learning paradigms and provides a robust benchmark for evaluating models in both Seen and Unseen scenarios. 

Note that PIAD does not include language annotations. Since PIAD and LASO share the same object classes, affordance categories, and the same $58$ affordance-object pairings, we reuse LASO's language annotations for PIAD. For each object and affordance category label in PIAD, we randomly select a question from LASO's question dataset corresponding to that affordance-object pairing.

\subsection{Evaluation Metrics}

To comprehensively evaluate the performance of our method, we employ four widely used metrics: \textbf{AUC}, \textbf{aIoU}, \textbf{SIM}, and \textbf{MAE}. Each metric is designed to assess different aspects of affordance prediction, providing a robust and multi-faceted evaluation framework. Below, we detail the formulation and significance of each metric:

\begin{itemize}
    \item \textbf{Area Under the ROC Curve (AUC)} \cite{lobo2008auc}:  
    AUC measures the model's ability to distinguish between regions of high and low affordance saliency on the point cloud. Specifically, the saliency map is treated as a binary classifier at various threshold levels, and a Receiver Operating Characteristic (ROC) curve is generated by plotting the true positive rate (TPR) against the false positive rate (FPR) at each threshold. AUC provides a single scalar value summarizing the overall performance, where higher values indicate better discrimination ability. It is particularly useful for comparing models' effectiveness in highlighting affordance-relevant regions.

    \item \textbf{Average Intersection over Union (aIoU)} \cite{rahman2016optimizing}:  
    IoU is a standard metric for comparing the similarity between two arbitrary regions—in this case, the predicted affordance region and the ground truth. It is defined as the size of the intersection between the two regions divided by the size of their union:
    \begin{equation}
        IoU = \frac{TP}{TP+FP+FN},
    \end{equation}
    where $TP$, $FP$, and $FN$ denote true positives, false positives, and false negatives, respectively. The aIoU extends this metric to compute the average IoU across all categories and test samples, providing a quantitative measure of the overlap between predicted and labeled affordance regions. Higher values indicate better alignment between the prediction and the ground truth.

    \item \textbf{Similarity (SIM)} \cite{swain1991color}:  
    The SIM metric evaluates how closely the predicted affordance map matches the ground truth by summing the minimum values at each point. For normalized prediction and ground truth maps $P$ and $Q$, the similarity is calculated as:
    \begin{equation}
        SIM(P,Q) = \sum_{i} \min(P_i, Q_i),
    \end{equation}
    where the inputs are normalized such that $\sum_{i}P_i = \sum_{i}Q_i = 1$. SIM provides a measure of how well the model captures the relative affordance distribution across the point cloud. A higher similarity score reflects greater consistency between the predicted and true maps, making it a valuable metric for evaluating spatial prediction fidelity.

    \item \textbf{Mean Absolute Error (MAE)} \cite{willmott2005advantages}:  
    MAE quantifies the average absolute difference between the predicted affordance values and the ground truth, offering a direct measure of prediction accuracy. For $n$ points in a point cloud, it is calculated as:
    \begin{equation}
        \mathrm{MAE} = \frac{1}{n} \sum_{i=1}^n \left| e_i \right|,
    \end{equation}
    where $e_i$ is the point-wise error. MAE is particularly useful for evaluating overall prediction quality by penalizing larger deviations. Lower MAE values indicate better performance, as they reflect a smaller error margin between the predicted and ground truth affordance scores.
\end{itemize}

\noindent Together, these metrics provide a comprehensive framework to benchmark the performance of affordance prediction models. AUC evaluates ranking capability, aIoU measures spatial overlap, SIM assesses prediction similarity, and MAE quantifies overall prediction accuracy. By combining these complementary metrics, we ensure a holistic evaluation of model performance under diverse scenarios.

\subsection{Baselines}

We evaluate our method against state-of-the-art approaches on both the PIAD and LASO datasets. Among these, LASO~\cite{li2024laso} is the closest to our method, as it also generates affordance scores based on textual cues. Additionally, we include 3D cross-modal baselines such as 3D-SPS~\cite{luo20223d}, and image segmentation methods like ReferTrans~\cite{li2021referring} and RelA~\cite{liu2023gres}, which leverage vision-language models for cross-modal alignment. Results for these methods are referenced directly from the LASO paper.

On the PIAD dataset, we compare against IAGNet~\cite{yang2023grounding}, a method that grounds 3D affordances by transferring knowledge from demonstration images into point clouds. Furthermore, this benchmark includes advanced image-point cloud cross-modal methods, including MBDF~\cite{tan2021mbdf}, PMF~\cite{zhuang2021perception}, FRCNN~\cite{xu2022fusionrcnn}, ILN~\cite{chen2022imlovenet}, PFusion~\cite{xu2018pointfusion}, and XMF~\cite{aiello2022cross}. These baselines align image and point cloud features in various ways. Results for these baselines are taken from the IAGNet paper, except for LASO, which is retrained in the PIAD setting.

Below is a brief introduction to the baselines:

\begin{itemize}

\item \textbf{LASO}~\cite{li2024laso}: Generates affordance segmentation masks using textual-conditioned affordance queries, focusing on cross-modal alignment between text and 3D objects.

\item \textbf{IAGNet}~\cite{yang2023grounding}: Grounds 3D affordances by transferring knowledge from 2D demonstration images to point clouds, leveraging cross-modal affordance reasoning.

\item \textbf{3D-SPS}~\cite{luo20223d}: A 3D visual grounding method that selects linguistic keypoints for affordance segmentation, adapted by removing its bounding box prediction module.

\item \textbf{ReLA}~\cite{liu2023gres}: Originally designed for image-based referring expression segmentation, it segments point clouds based on language expressions by adapting image region features to grouped point features.

\item \textbf{ReferTrans}~\cite{li2021referring}: A transformer-based architecture for image-based expression segmentation, modified for point clouds by replacing the image backbone with a 3D backbone and focusing solely on mask prediction.

\item \textbf{MBDF-Net (MBDF)}~\cite{tan2021mbdf}: Employs an Adaptive Attention Fusion (AAF) module for cross-modal feature fusion, with modifications to exclude camera intrinsic parameters.

\item \textbf{PMF}~\cite{zhuang2021perception}: Uses a residual-based fusion model to combine image and point cloud features, incorporating convolution and attention, while omitting perspective projection.

\item \textbf{FusionRCNN (FRCNN)}~\cite{xu2022fusionrcnn}: Fuses proposals extracted from images and point clouds through iterative self-attention and cross-attention mechanisms.

\item \textbf{ImloveNet (ILN)}~\cite{chen2022imlovenet}: Projects image features into 3D space using a learnable mapping, and fuses these with point cloud features using an attention mechanism.

\item \textbf{PointFusion (PFusion)}~\cite{xu2018pointfusion}: Performs dense fusion by combining global and point-wise features extracted separately from point clouds and images.

\item \textbf{XMFnet (XMF)}~\cite{aiello2022cross}: Fuses localized features from point clouds and images using a combination of cross-attention and self-attention, originally designed for cross-modal point cloud completion.
\end{itemize}

\section{Additional Quantitative Results}
In this section, we provide additional quantitative results, \ie, the class-wise and corruption-wise evaluation metrics, to demonstrate the effectiveness of our method.

\begin{table*}[t]
    \centering
    \caption{The category-wise results for LASO~\cite{li2024laso} and GEAL (Ours) on the \textbf{Seen} partition of the \textbf{PIAD} dataset~\cite{yang2023grounding}. AUC and aIOU scores are reported in percentages (\%).}
    \vspace{-0.2cm}
    \resizebox{0.8\linewidth}{!}{
    \begin{tabular}{c|r|cccc|cccc}
    \toprule
    & & \multicolumn{4}{c|}{\textbf{LASO~\cite{li2024laso}}} & \multicolumn{4}{c}{\textbf{GEAL (Ours)}} \\
    \textbf{\#} & \textbf{Category} & \textbf{aIOU} $\uparrow$ & \textbf{AUC} $\uparrow$ & \textbf{SIM} $\uparrow$ & \textbf{MAE} $\downarrow$ & \textbf{aIOU} $\uparrow$ & \textbf{AUC} $\uparrow$ & \textbf{SIM} $\uparrow$ & \textbf{MAE} $\downarrow$ \\
    \midrule
    \midrule
    $1$ & \textbf{Bag} \textcolor{BurntOrange}{$\bullet$} & $23.4$ & $83.3$ & $0.567$ & $0.090$ & $24.0$ & $85.1$ & $0.588$ & $0.088$ \\
    \rowcolor{gray!10}$2$ & \textbf{Bed} \textcolor{BurntOrange}{$\bullet$} & $21.1$ & $87.3$ & $0.587$ & $0.097$ & $22.7$ & $88.1$ & $0.595$ & $0.091$ \\
    $3$ & \textbf{Bowl} \textcolor{BurntOrange}{$\bullet$} & $7.4$ & $76.2$ & $0.736$ & $0.114$ & $9.8$ & $84.1$ & $0.761$ & $0.105$ \\
    \rowcolor{gray!10}$4$ & \textbf{Clock} \textcolor{BurntOrange}{$\bullet$} & $7.5$ & $91.5$ & $0.473$ & $0.077$ & $11.1$ & $92.5$ & $0.596$ & $0.051$ \\
    $5$ & \textbf{Dishwash} \textcolor{BurntOrange}{$\bullet$} & $24.7$ & $91.9$ & $0.464$ & $0.069$ & $26.2$ & $92.9$ & $0.496$ & $0.058$ \\
    \rowcolor{gray!10}$6$ & \textbf{Display} \textcolor{BurntOrange}{$\bullet$} & $32.5$ & $91.5$ & $0.719$ & $0.083$ & $37.7$ & $91.3$ & $0.726$ & $0.104$ \\
    $7$ & \textbf{Door} \textcolor{BurntOrange}{$\bullet$} & $10.1$ & $81.2$ & $0.437$ & $0.064$ & $11.0$ & $83.8$ & $0.395$ & $0.054$ \\
    \rowcolor{gray!10}$8$ & \textbf{Earphone} \textcolor{BurntOrange}{$\bullet$} & $18.8$ & $85.9$ & $0.615$ & $0.094$ & $21.6$ & $87.6$ & $0.654$ & $0.086$ \\
    $9$ & \textbf{Faucet} \textcolor{BurntOrange}{$\bullet$} & $19.9$ & $79.9$ & $0.517$ & $0.099$ & $19.1$ & $83.6$ & $0.602$ & $0.078$ \\
    \rowcolor{gray!10}$10$ & \textbf{Hat} \textcolor{BurntOrange}{$\bullet$} & $4.7$ & $65.9$ & $0.604$ & $0.148$ & $7.8$ & $74.2$ & $0.620$ & $0.133$ \\
    $11$ & ~\textbf{StorageFurniture} \textcolor{BurntOrange}{$\bullet$} & $17.3$ & $87.2$ & $0.419$ & $0.077$ & $20.8$ & $87.5$ & $0.430$ & $0.065$ \\
    \rowcolor{gray!10}$12$ & \textbf{Keyboard} \textcolor{BurntOrange}{$\bullet$} & $14.8$ & $81.2$ & $0.249$ & $0.059$ & $15.2$ & $84.6$ & $0.257$ & $0.048$ \\
    $13$ & \textbf{Knife} \textcolor{BurntOrange}{$\bullet$} & $15.5$ & $89.8$ & $0.671$ & $0.060$ & $23.5$ & $94.1$ & $0.717$ & $0.046$ \\
    \rowcolor{gray!10}$14$ & \textbf{Laptop} \textcolor{BurntOrange}{$\bullet$} & $29.2$ & $94.1$ & $0.566$ & $0.072$ & $31.2$ & $94.2$ & $0.575$ & $0.069$ \\
    $15$ & \textbf{Microwave} \textcolor{BurntOrange}{$\bullet$} & $30.1$ & $96.8$ & $0.524$ & $0.037$ & $35.5$ & $96.9$ & $0.545$ & $0.037$ \\
    \rowcolor{gray!10}$16$ & \textbf{Mug} \textcolor{BurntOrange}{$\bullet$} & $10.7$ & $76.5$ & $0.578$ & $0.107$ & $17.5$ & $77.2$ & $0.607$ & $0.091$ \\
    $17$ & \textbf{Refrigerator} \textcolor{BurntOrange}{$\bullet$} & $23.2$ & $87.1$ & $0.473$ & $0.070$ & $24.7$ & $89.6$ & $0.460$ & $0.070$ \\
    \rowcolor{gray!10}$18$ & \textbf{Chair} \textcolor{BurntOrange}{$\bullet$} & $27.5$ & $88.1$ & $0.649$ & $0.094$ & $28.5$ & $89.0$ & $0.652$ & $0.066$ \\
    $19$ & \textbf{Scissors} \textcolor{BurntOrange}{$\bullet$} & $24.1$ & $91.2$ & $0.631$ & $0.055$ & $31.9$ & $95.8$ & $0.698$ & $0.040$ \\
    \rowcolor{gray!10}$20$ & \textbf{Table} \textcolor{BurntOrange}{$\bullet$} & $10.1$ & $78.2$ & $0.627$ & $0.129$ & $11.4$ & $79.1$ & $0.639$ & $0.135$ \\
    $21$ & \textbf{TrashCan} \textcolor{BurntOrange}{$\bullet$} & $11.9$ & $67.4$ & $0.323$ & $0.143$ & $16.2$ & $68.8$ & $0.385$ & $0.146$ \\
    \rowcolor{gray!10}$22$ & \textbf{Vase} \textcolor{BurntOrange}{$\bullet$} & $10.3$ & $72.0$ & $0.608$ & $0.120$ & $12.5$ & $72.4$ & $0.612$ & $0.116$ \\
    $23$ & \textbf{Bottle} \textcolor{BurntOrange}{$\bullet$} & $23.5$ & $77.3$ & $0.552$ & $0.110$ & $27.8$ & $79.8$ & $0.536$ & $0.107$ \\
    \bottomrule
    \end{tabular}
    }
\label{tab:append_piad_seen}
\end{table*}
\begin{table*}[t]
    \centering
    \caption{The category-wise results for LASO~\cite{li2024laso} and GEAL (Ours) on the \textbf{Unseen} partition of the \textbf{PIAD} dataset~\cite{yang2023grounding}. AUC and aIOU scores are reported in percentages (\%).}
    \vspace{-0.2cm}
    \resizebox{0.8\linewidth}{!}{
    \begin{tabular}{c|r|cccc|cccc}
    \toprule
    & & \multicolumn{4}{c|}{\textbf{LASO~\cite{li2024laso}}} & \multicolumn{4}{c}{\textbf{GEAL (Ours)}} \\
    \textbf{\#} & \textbf{Category} & \textbf{aIOU} $\uparrow$ & \textbf{AUC} $\uparrow$ & \textbf{SIM} $\uparrow$ & \textbf{MAE} $\downarrow$ & \textbf{aIOU} $\uparrow$ & \textbf{AUC} $\uparrow$ & \textbf{SIM} $\uparrow$ & \textbf{MAE} $\downarrow$ \\
    \midrule
    \midrule
    $1$ & \textbf{Bed} \textcolor{BurntOrange}{$\bullet$} & $12.0$ & $78.0$ & $0.469$ & $0.126$ & $12.8$ & $78.4$ & $0.473$ & $0.120$ \\
    \rowcolor{gray!10}$2$ & \textbf{Dishwasher} \textcolor{BurntOrange}{$\bullet$} & $17.3$ & $84.9$ & $0.338$ & $0.079$ & $18.3$ & $89.8$ & $0.440$ & $0.060$ \\
    $3$ & \textbf{Laptop} \textcolor{BurntOrange}{$\bullet$} & $4.5$ & $65.4$ & $0.192$ & $0.122$ & $6.3$ & $74.5$ & $0.201$ & $0.100$ \\
    \rowcolor{gray!10}$4$ & \textbf{Microwave} \textcolor{BurntOrange}{$\bullet$} & $14.4$ & $83.4$ & $0.365$ & $0.066$ & $15.8$ & $89.6$ & $0.402$ & $0.049$ \\
    $5$ & \textbf{Scissors} \textcolor{BurntOrange}{$\bullet$} & $3.2$ & $66.5$ & $0.310$ & $0.107$ & $3.7$ & $69.8$ & $0.333$ & $0.123$ \\
    \rowcolor{gray!10}$6$ & \textbf{Vase} \textcolor{BurntOrange}{$\bullet$} & $5.2$ & $58.1$ & $0.455$ & $0.140$ & $6.4$ & $54.9$ & $0.466$ & $0.127$ \\
    \bottomrule
    \end{tabular}
    }
    \label{tab:append_piad_unseen}
\end{table*}

\begin{table*}[t]
    \centering
    \caption{The category-wise results for LASO~\cite{li2024laso} and GEAL (Ours) on the \textbf{Seen} partition of the \textbf{LASO} dataset~\cite{li2024laso}. AUC and aIOU scores are reported in percentages (\%).}
    \vspace{-0.2cm}
    \resizebox{0.8\linewidth}{!}{
    \begin{tabular}{c|r|cccc|cccc}
    \toprule
    & & \multicolumn{4}{c|}{\textbf{LASO~\cite{li2024laso}}} & \multicolumn{4}{c}{\textbf{GEAL (Ours)}} \\
    \textbf{\#} & \textbf{Category} & \textbf{aIOU} $\uparrow$ & \textbf{AUC} $\uparrow$ & \textbf{SIM} $\uparrow$ & \textbf{MAE} $\downarrow$ & \textbf{aIOU} $\uparrow$ & \textbf{AUC} $\uparrow$ & \textbf{SIM} $\uparrow$ & \textbf{MAE} $\downarrow$ \\
    \midrule
    \midrule
    $1$ & \textbf{Bag} \textcolor{Emerald}{$\bullet$} & $19.8$ & $85.4$ & $0.535$ & $0.085$ & $20.6$ & $86.7$ & $0.572$ & $0.084$ \\
    \rowcolor{gray!10}$2$ & \textbf{Bed} \textcolor{Emerald}{$\bullet$} & $13.6$ & $77.4$ & $0.515$ & $0.111$ & $16.0$ & $79.9$ & $0.527$ & $0.110$ \\
    $3$ & \textbf{Bowl} \textcolor{Emerald}{$\bullet$} & $8.6$ & $81.3$ & $0.777$ & $0.102$ & $12.2$ & $87.4$ & $0.807$ & $0.102$ \\
    \rowcolor{gray!10}$4$ & \textbf{Clock} \textcolor{Emerald}{$\bullet$} & $6.2$ & $84.2$ & $0.461$ & $0.064$ & $9.8$ & $84.8$ & $0.485$ & $0.062$ \\
    $5$ & \textbf{Dishwash} \textcolor{Emerald}{$\bullet$} & $29.6$ & $94.1$ & $0.472$ & $0.070$ & $28.5$ & $89.9$ & $0.505$ & $0.068$ \\
    \rowcolor{gray!10}$6$ & \textbf{Display} \textcolor{Emerald}{$\bullet$} & $31.0$ & $92.2$ & $0.700$ & $0.086$ & $41.1$ & $92.6$ & $0.718$ & $0.088$ \\
    $7$ & \textbf{Door} \textcolor{Emerald}{$\bullet$} & $12.3$ & $82.3$ & $0.311$ & $0.060$ & $15.7$ & $83.8$ & $0.368$ & $0.058$ \\
    \rowcolor{gray!10}$8$ & \textbf{Earphone} \textcolor{Emerald}{$\bullet$} & $26.5$ & $93.0$ & $0.639$ & $0.099$ & $27.5$ & $94.0$ & $0.662$ & $0.094$ \\
    $9$ & \textbf{Faucet} \textcolor{Emerald}{$\bullet$} & $14.2$ & $78.9$ & $0.498$ & $0.089$ & $18.3$ & $84.3$ & $0.589$ & $0.087$ \\
    \rowcolor{gray!10}$10$ & \textbf{Hat} \textcolor{Emerald}{$\bullet$} & $3.6$ & $67.0$ & $0.538$ & $0.152$ & $9.3$ & $72.7$ & $0.560$ & $0.148$ \\
    $11$ & ~\textbf{StorageFurniture} \textcolor{Emerald}{$\bullet$} & $19.2$ & $88.6$ & $0.437$ & $0.067$ & $24.7$ & $89.3$ & $0.481$ & $0.066$ \\
    \rowcolor{gray!10}$12$ & \textbf{Keyboard} \textcolor{Emerald}{$\bullet$} & $12.0$ & $89.0$ & $0.227$ & $0.055$ & $12.9$ & $87.9$ & $0.232$ & $0.039$ \\
    $13$ & \textbf{Knife} \textcolor{Emerald}{$\bullet$} & $14.8$ & $91.3$ & $0.642$ & $0.064$ & $22.9$ & $93.2$ & $0.657$ & $0.063$ \\
    \rowcolor{gray!10}$14$ & \textbf{Laptop} \textcolor{Emerald}{$\bullet$} & $28.5$ & $95.1$ & $0.583$ & $0.078$ & $29.8$ & $95.1$ & $0.586$ & $0.070$ \\
    $15$ & \textbf{Microwave} \textcolor{Emerald}{$\bullet$} & $27.2$ & $96.1$ & $0.440$ & $0.042$ & $31.8$ & $92.8$ & $0.464$ & $0.038$ \\
    \rowcolor{gray!10}$16$ & \textbf{Mug} \textcolor{Emerald}{$\bullet$} & $13.3$ & $78.1$ & $0.547$ & $0.098$ & $21.7$ & $87.6$ & $0.635$ & $0.076$ \\
    $17$ & \textbf{Refrigerator} \textcolor{Emerald}{$\bullet$} & $25.6$ & $92.8$ & $0.433$ & $0.063$ & $24.8$ & $93.7$ & $0.484$ & $0.069$ \\
    \rowcolor{gray!10}$18$ & \textbf{Chair} \textcolor{Emerald}{$\bullet$} & $28.9$ & $89.9$ & $0.650$ & $0.093$ & $28.7$ & $89.9$ & $0.678$ & $0.091$ \\
    $19$ & \textbf{Scissors} \textcolor{Emerald}{$\bullet$} & $17.5$ & $95.4$ & $0.661$ & $0.053$ & $24.9$ & $95.9$ & $0.684$ & $0.045$ \\
    \rowcolor{gray!10}$20$ & \textbf{Table} \textcolor{Emerald}{$\bullet$} & $10.1$ & $81.7$ & $0.662$ & $0.119$ & $10.8$ & $81.6$ & $0.690$ & $0.115$ \\
    $21$ & \textbf{TrashCan} \textcolor{Emerald}{$\bullet$} & $10.9$ & $72.1$ & $0.323$ & $0.137$ & $27.8$ & $90.4$ & $0.499$ & $0.100$ \\
    \rowcolor{gray!10}$22$ & \textbf{Vase} \textcolor{Emerald}{$\bullet$} & $7.9$ & $71.1$ & $0.630$ & $0.125$ & $13.5$ & $79.5$ & $0.650$ & $0.116$ \\
    $23$ & \textbf{Bottle} \textcolor{Emerald}{$\bullet$} & $20.4$ & $81.2$ & $0.553$ & $0.114$ & $28.7$ & $81.9$ & $0.570$ & $0.116$ \\
    \bottomrule
    \end{tabular}
    }
\label{tab:append_laso_seen}
\end{table*}

\begin{table*}[t]
    \centering
    \caption{The category-wise results for LASO~\cite{li2024laso} and GEAL (Ours) on the \textbf{Unseen} partition of the \textbf{LASO} dataset~\cite{li2024laso}. AUC and aIOU scores are reported in percentages (\%).}
    \vspace{-0.2cm}
    \resizebox{0.8\linewidth}{!}{
    \begin{tabular}{c|r|cccc|cccc}
    \toprule
    & & \multicolumn{4}{c|}{\textbf{LASO~\cite{li2024laso}}} & \multicolumn{4}{c}{\textbf{GEAL (Ours)}} \\
    \textbf{\#} & \textbf{Category} & \textbf{aIOU} $\uparrow$ & \textbf{AUC} $\uparrow$ & \textbf{SIM} $\uparrow$ & \textbf{MAE} $\downarrow$ & \textbf{aIOU} $\uparrow$ & \textbf{AUC} $\uparrow$ & \textbf{SIM} $\uparrow$ & \textbf{MAE} $\downarrow$ \\
    \midrule
    \midrule
    $1$ & \textbf{Bag} \textcolor{Emerald}{$\bullet$} & $20.7$ & $89.1$ & $0.513$ & $0.089$ & $22.1$ & $91.0$ & $0.522$ & $0.086$ \\
    \rowcolor{gray!10}$2$ & \textbf{Bed} \textcolor{Emerald}{$\bullet$} & $12.2$ & $80.6$ & $0.553$ & $0.115$ & $13.6$ & $81.4$ & $0.563$ & $0.113$ \\
    $3$ & \textbf{Bowl} \textcolor{Emerald}{$\bullet$} & $7.5$ & $81.3$ & $0.744$ & $0.125$ & $9.1$ & $82.5$ & $0.749$ & $0.119$ \\
    \rowcolor{gray!10}$4$ & \textbf{Clock} \textcolor{Emerald}{$\bullet$} & $5.3$ & $85.2$ & $0.419$ & $0.094$ & $6.4$ & $85.0$ & $0.433$ & $0.079$ \\
    $5$ & \textbf{Dishwash} \textcolor{Emerald}{$\bullet$} & $20.7$ & $92.4$ & $0.443$ & $0.069$ & $26.0$ & $92.4$ & $0.470$ & $0.065$ \\
    \rowcolor{gray!10}$6$ & \textbf{Display} \textcolor{Emerald}{$\bullet$} & $23.4$ & $86.6$ & $0.512$ & $0.112$ & $25.0$ & $87.6$ & $0.526$ & $0.112$ \\
    $7$ & \textbf{Door} \textcolor{Emerald}{$\bullet$} & $3.4$ & $81.3$ & $0.324$ & $0.095$ & $11.7$ & $81.4$ & $0.355$ & $0.066$ \\
    \rowcolor{gray!10}$8$ & \textbf{Earphone} \textcolor{Emerald}{$\bullet$} & $9.5$ & $76.8$ & $0.454$ & $0.130$ & $20.8$ & $93.5$ & $0.639$ & $0.091$ \\
    $9$ & \textbf{Faucet} \textcolor{Emerald}{$\bullet$} & $13.8$ & $74.1$ & $0.442$ & $0.098$ & $15.1$ & $76.8$ & $0.470$ & $0.095$ \\
    \rowcolor{gray!10}$10$ & \textbf{Hat} \textcolor{Emerald}{$\bullet$} & $4.5$ & $61.2$ & $0.586$ & $0.158$ & $4.1$ & $66.5$ & $0.582$ & $0.149$ \\
    $11$ & \textbf{StorageFurniture} \textcolor{Emerald}{$\bullet$} & $17.9$ & $88.1$ & $0.422$ & $0.069$ & $18.3$ & $88.3$ & $0.423$ & $0.067$ \\
    \rowcolor{gray!10}$12$ & \textbf{Keyboard} \textcolor{Emerald}{$\bullet$} & $3.1$ & $74.6$ & $0.138$ & $0.082$ & $3.3$ & $79.4$ & $0.137$ & $0.078$ \\
    $13$ & \textbf{Knife} \textcolor{Emerald}{$\bullet$} & $15.3$ & $91.7$ & $0.643$ & $0.053$ & $15.4$ & $91.2$ & $0.675$ & $0.059$ \\
    \rowcolor{gray!10}$14$ & \textbf{Laptop} \textcolor{Emerald}{$\bullet$} & $8.7$ & $79.7$ & $0.334$ & $0.096$ & $29.3$ & $95.6$ & $0.610$ & $0.064$ \\
    $15$ & \textbf{Microwave} \textcolor{Emerald}{$\bullet$} & $11.9$ & $90.9$ & $0.317$ & $0.063$ & $14.2$ & $91.5$ & $0.318$ & $0.064$ \\
    \rowcolor{gray!10}$16$ & \textbf{Mug} \textcolor{Emerald}{$\bullet$} & $1.7$ & $64.5$ & $0.381$ & $0.174$ & $2.5$ & $66.6$ & $0.511$ & $0.157$ \\
    $17$ & \textbf{Refrigerator} \textcolor{Emerald}{$\bullet$} & $20.1$ & $87.2$ & $0.378$ & $0.066$ & $21.0$ & $89.4$ & $0.390$ & $0.065$ \\
    \rowcolor{gray!10}$18$ & \textbf{Chair} \textcolor{Emerald}{$\bullet$} & $25.2$ & $87.4$ & $0.642$ & $0.098$ & $26.0$ & $89.4$ & $0.624$ & $0.094$ \\
    $19$ & \textbf{Scissors} \textcolor{Emerald}{$\bullet$} & $1.6$ & $25.3$ & $0.094$ & $0.105$ & $2.1$ & $27.6$ & $0.105$ & $0.097$ \\
    \rowcolor{gray!10}$20$ & \textbf{Table} \textcolor{Emerald}{$\bullet$} & $7.5$ & $70.4$ & $0.604$ & $0.135$ & $7.8$ & $72.1$ & $0.620$ & $0.129$ \\
    $21$ & \textbf{TrashCan} \textcolor{Emerald}{$\bullet$} & $2.6$ & $63.1$ & $0.191$ & $0.124$ & $7.4$ & $71.0$ & $0.293$ & $0.125$ \\
    \rowcolor{gray!10}$22$ & \textbf{Vase} \textcolor{Emerald}{$\bullet$} & $6.4$ & $56.4$ & $0.466$ & $0.148$ & $7.6$ & $67.0$ & $0.614$ & $0.140$ \\
    $23$ & \textbf{Bottle} \textcolor{Emerald}{$\bullet$} & $16.2$ & $78.5$ & $0.455$ & $0.134$ & $21.2$ & $78.2$ & $0.519$ & $0.119$ \\
    \bottomrule
    \end{tabular}
    }
\label{tab:append_laso_unseen}
\end{table*}

\subsection{Complete Results on PIAD}

The complete results of the comparative methods for all object categories in the \textbf{Seen} and \textbf{Unseen} partitions of the PIAD dataset~\cite{yang2023grounding} are provided in~\cref{tab:append_piad_seen} and~\cref{tab:append_piad_unseen}, respectively.

\subsection{Complete Results on LASO}

The complete results of the comparative methods for all object categories in the \textbf{Seen} and \textbf{Unseen} partitions of the LASO dataset~\cite{li2024laso} are provided in~\cref{tab:append_laso_seen} and~\cref{tab:append_laso_unseen}, respectively.

\section{Additional Qualitative Results}
In this section, we provide more qualitative results (visual examples) to demonstrate the effectiveness of our method.

\subsection{Additional Qualitative Results on PIAD-C}
\begin{figure*}[t]
    \centering
    \includegraphics[width=1\linewidth]{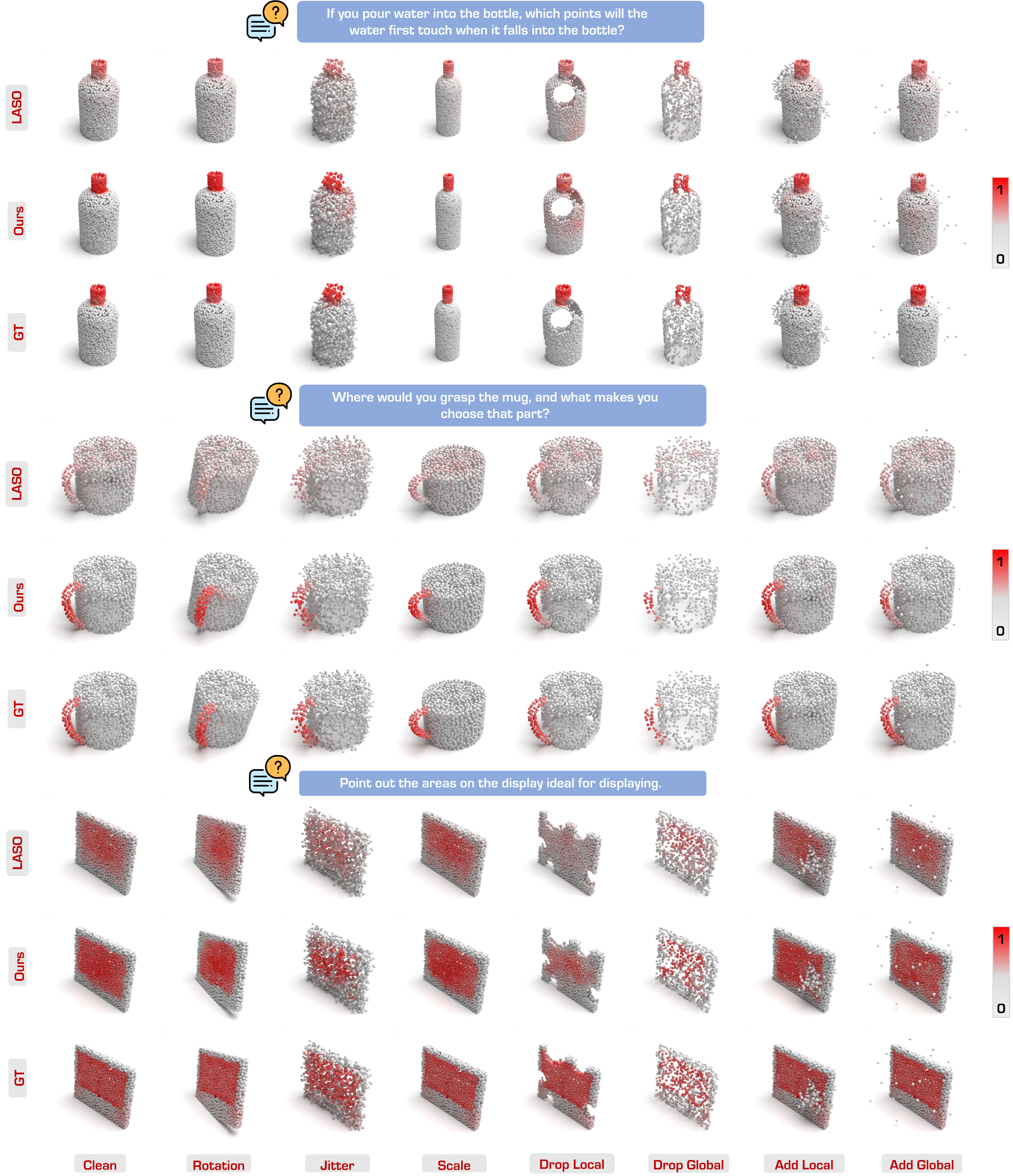}
    \caption{Qualitative comparisons between \textbf{GEAL} and LASO~\cite{li2024laso} on the PIAD-C dataset, highlighting the superior robustness of our method on corrupted data. 
    }
  \label{fig:append_corrupt}
\end{figure*}

We include additional qualitative results of \textbf{GEAL} and LASO~\cite{li2024laso} on the PIAD-C dataset in ~\cref{fig:append_corrupt}.

\subsection{Additional Qualitative Results on PIAD}
\begin{figure*}[t]
    \centering
    \includegraphics[width=\linewidth]{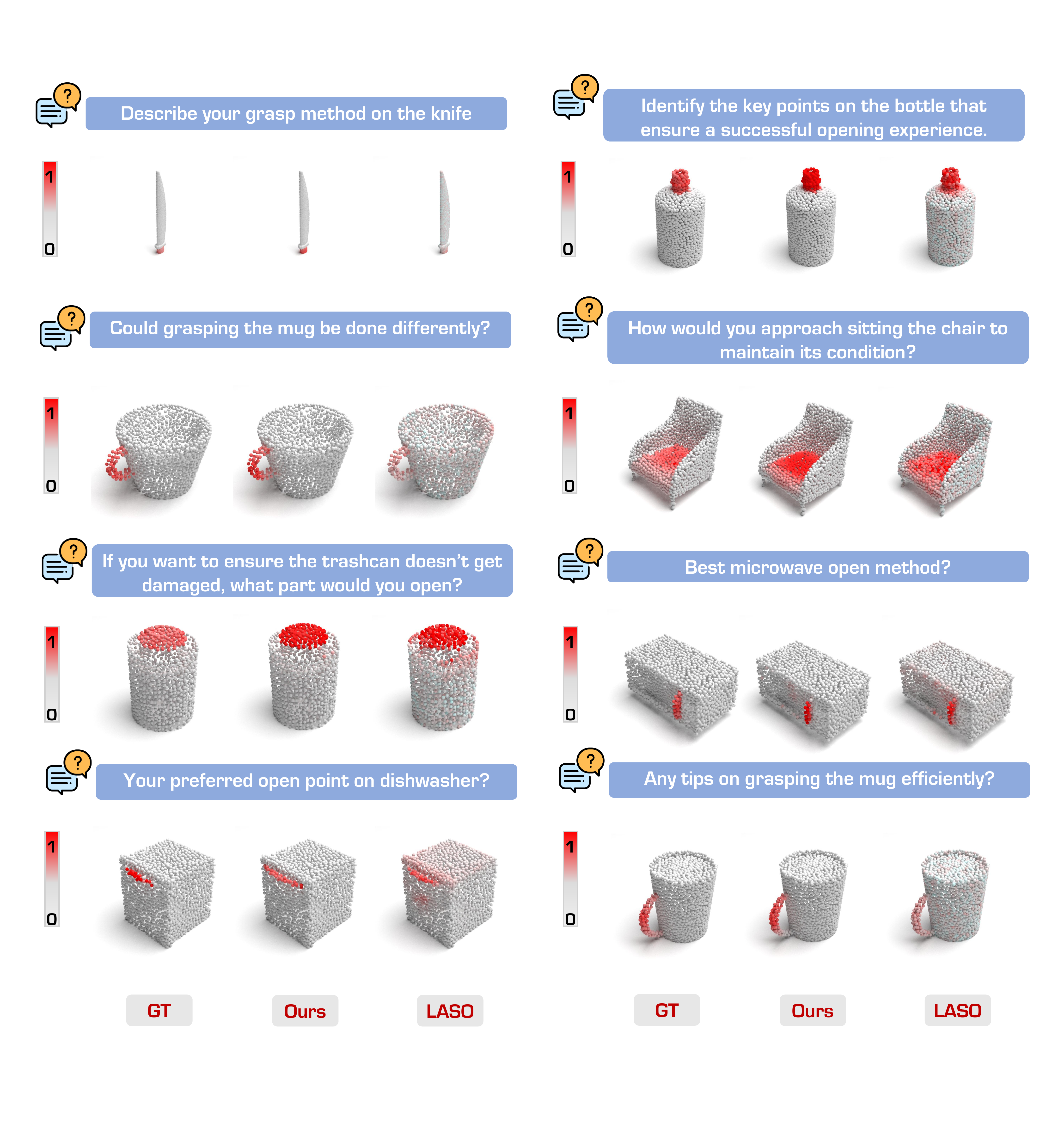}
    \vspace{-2cm}
    \caption{Qualitative comparisons between \textbf{GEAL} and LASO~\cite{li2024laso} on the PIAD dataset.
    }
    \label{fig:append_seen}
\end{figure*}

We include additional qualitative results of \textbf{GEAL} and LASO~\cite{li2024laso} on the PIAD dataset in ~\cref{fig:append_seen}.

\clearpage
\clearpage
\section{Broader Impact \& Limitations}
\label{sec:limitation}
In this section, we discuss the societal impact, broader impact, and potential limitations of this work.

\subsection{Societal Impact}
The proposed framework for 3D affordance learning has significant societal implications, enabling embodied intelligence for effective robot and AI interaction with surroundings. This advancement can enhance automated systems' efficiency and safety in fields like healthcare, elderly care, and disaster response, where understanding object affordances is critical. This technology also has the potential to empower individuals with disabilities by enabling assistive robots to perform tasks such as fetching, opening, or manipulating objects. Applications in education and augmented or virtual reality could transform learning and entertainment by offering immersive and interactive experiences.

\subsection{Broader Impact}
Affordance learning can redefine robotics automation by improving autonomy and adaptability in industries. In manufacturing, it allows robots to handle diverse objects with minimal reprogramming, optimizing workflows and reducing human workload. In agriculture and environmental monitoring, affordance-aware systems can adapt to dynamic environments for precise operations. Integrating affordance grounding with augmented and virtual reality enables new possibilities in training, simulation, and interactive applications. This could drive innovations in medical training, such as AR-guided surgeries, and in gaming, offering intuitive and immersive user experiences through affordance-based interactions.

\subsection{Potential Limitations}
Despite its advantages, the proposed framework may encounter certain limitations:

\begin{itemize}
\item \textbf{Limited Generalization for Internal Affordances:} The framework struggles to accurately perceive and generalize affordances associated with the internal properties of objects, such as the "contain" affordance of a bottle. This limitation arises because point cloud processing primarily focuses on an object's external surface, often neglecting internal structures. Furthermore, the scarcity of high-quality data representing internal affordances, hampers the system's ability to generalize on such affordances.

\item \textbf{Ethical Concerns:} In applications such as surveillance or autonomous decision-making, the deployment of the framework introduces potential ethical concerns. Misuse of the technology could infringe on privacy or lead to a lack of accountability in critical decision-making scenarios, highlighting the importance of establishing robust ethical guidelines for its use.

\item \textbf{Resource Intensity:} Training and deploying such sophisticated models demand significant computational resources, which can pose a challenge for smaller organizations or regions with limited access to advanced technology infrastructure. This barrier may restrict the broader adoption of the framework in resource-constrained environments.
\end{itemize}

\section{Public Resource Used}
\label{sec:ack}
In this section, we acknowledge the use of the following public resources, during the course of this work:

\begin{itemize}
    \item LASO\footnote{\url{https://github.com/yl3800/LASO}}\dotfill Unknown
    \item IAGNet\footnote{\url{https://github.com/yyvhang/IAGNet}}\dotfill Unknown
    \item PointCloud-C\footnote{\url{https://github.com/ldkong1205/PointCloud-C}}\dotfill Unknown
    \item OOAL\footnote{\url{https://github.com/Reagan1311/OOAL}}\dotfill MIT License
    \item DreamGaussian\footnote{\url{https://github.com/dreamgaussian/dreamgaussian}}\dotfill MIT License
    \item LangSplat\footnote{\url{https://github.com/minghanqin/LangSplat}}\dotfill Gaussian-Splatting License
\end{itemize}
\clearpage
{
    \small
    \bibliographystyle{ieeenat_fullname}
    \bibliography{main}
}

\end{document}